\documentclass[10pt,twocolumn,letterpaper]{article}

\usepackage[pagenumbers]{iccv} %

\usepackage{amsmath,amsfonts,amsthm,amssymb}
\usepackage{mathtools}
\usepackage{bm}
\usepackage{nicefrac}
\usepackage{microtype}
\usepackage{lipsum}

\usepackage{color,xcolor}
\usepackage{epsfig}
\usepackage{graphicx}

\usepackage{adjustbox}
\usepackage{array}
\usepackage{booktabs}
\usepackage{colortbl}
\usepackage{wrapfig}
\usepackage{hhline}
\usepackage{multirow}
\usepackage{subcaption}
\usepackage[size=small]{caption}
\usepackage{float}

\usepackage{changepage}
\usepackage{extramarks}
\usepackage{fancyhdr}
\usepackage{lastpage}
\usepackage{setspace}
\usepackage{soul}
\usepackage{xspace}

\usepackage{url}
\usepackage{algorithm}
\usepackage{algorithmicx}
\usepackage{algpseudocode}
\usepackage{enumerate}
\usepackage{enumitem}  %
\usepackage{makecell}
\usepackage{pifont}
\usepackage{titlecaps}
\usepackage[accsupp]{axessibility}
\usepackage{framed}

\definecolor{Cardinal}{rgb}{0.549,0.082,0.082}

\newcommand{\model}{ZeroHSI\xspace}
\newcommand{\dataset}{AnyInteraction\xspace}

\newcommand{\myparagraph}[1]{%
    \vspace{0.1cm}
    \noindent
    \textbf{#1}%
}

\definecolor{MyDarkBlue}{rgb}{0,0.08,1}
\definecolor{MyAqua}{rgb}{0,0.7,0.7}
\definecolor{MyDarkGreen}{rgb}{0.02,0.6,0.02}
\definecolor{MyDarkRed}{rgb}{0.8,0.02,0.02}
\definecolor{MyDarkOrange}{rgb}{0.40,0.2,0.02}
\definecolor{MyPurple}{RGB}{111,0,255}
\definecolor{MyRed}{rgb}{1.0,0.0,0.0}
\definecolor{MyGold}{rgb}{0.75,0.6,0.12}
\definecolor{MyDarkgray}{rgb}{0.66, 0.66, 0.66}
\definecolor{MyGreen}{rgb}{0.00, 1.00, 0.68}

\definecolor{Cardinal}{rgb}{0.549,0.082,0.082}

\definecolor{iccvblue}{rgb}{0.21,0.49,0.74}

\definecolor{iccvblue}{rgb}{0.21,0.49,0.74}
\usepackage[pagebackref,breaklinks,colorlinks,allcolors=iccvblue]{hyperref}

\def\paperID{6501} %
\def\confName{ICCV}
\def\confYear{2025}

\title{\vspace{-1em}\model: Zero-Shot 4D Human-Scene Interaction by Video Generation\vspace{-9pt}}

\author{
    Hongjie Li\footnotemark[1] \qquad 
    Hong-Xing Yu\footnotemark[1] \qquad 
    Jiaman Li \qquad 
    Jiajun Wu \\
    \\
    Stanford University 
    \vspace{-21pt}
}

\begin{document}

\twocolumn[{%
\renewcommand\twocolumn[1][]{#1}%
\maketitle
\begin{center}
    \centering
    \vspace{8pt}
    \captionsetup{type=figure}
    \includegraphics[width=1\textwidth]{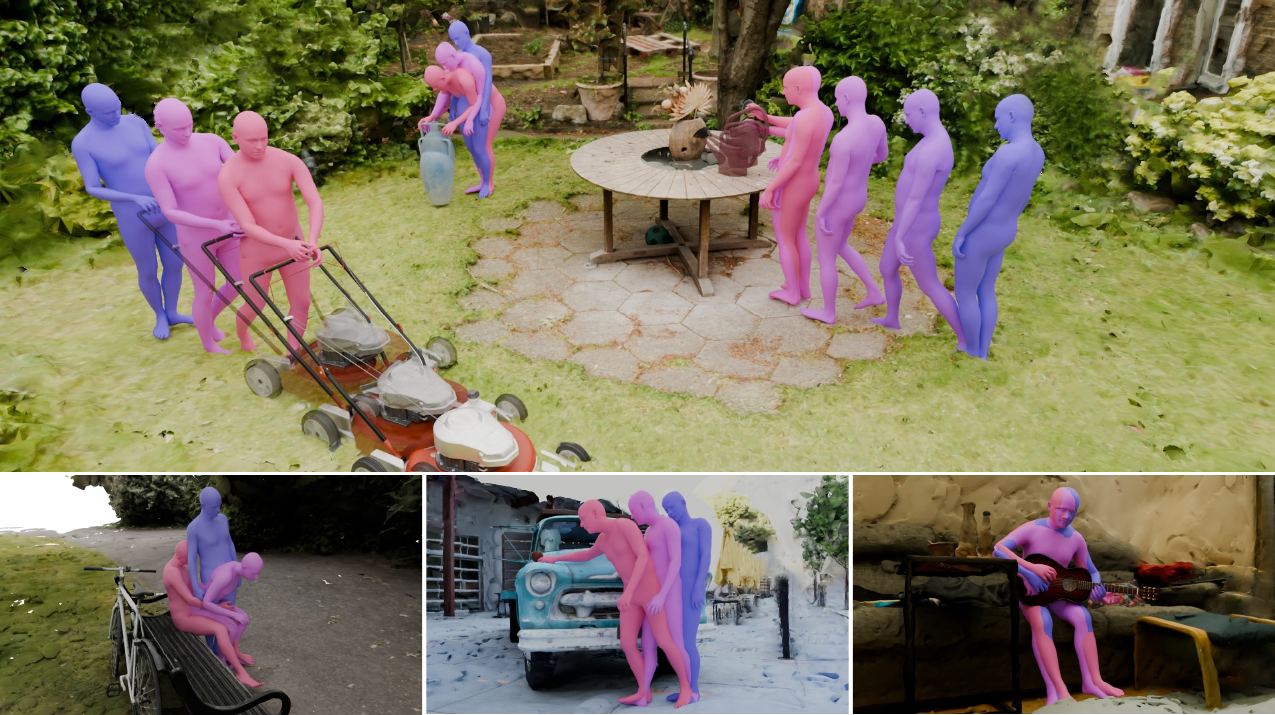}
    \caption{\textbf{Zero-shot human-scene interaction motion synthesis.} Our zero-shot method distills 4D interactions from video generation models to generate natural HSIs in various 3D environments. We demonstrate our method's effectiveness on real-world scenes reconstructed from the Mip-NeRF 360 (Garden, Bicycle, Room) and Tanks and Temples (Truck) datasets, showcasing diverse interactions with both static environments (walking, sitting, cleaning car) and dynamic objects (watering plants, lifting vase, operating mower, playing guitar). 
    }
    \label{fig:teaser}
\end{center}%
}]

\begin{abstract}
\vspace{-6mm}
\renewcommand{\thefootnote}{\fnsymbol{footnote}}\footnotetext[1]{Equal contribution. Work was done while H. Li was a visiting student at Stanford University. H. Li is now with Peking University.}\renewcommand{\thefootnote}{\arabic{footnote}}\setcounter{footnote}{0}

Human-scene interaction (HSI) generation is crucial for applications in embodied AI, virtual reality, and robotics. Yet, existing methods cannot synthesize interactions in unseen environments such as in-the-wild scenes or reconstructed scenes, as they rely on paired 3D scenes and captured human motion data for training, which are unavailable for unseen environments. We present \model, a novel approach that enables \textbf{zero-shot} 4D human-scene interaction synthesis, eliminating the need for training on any MoCap data. Our key insight is to distill human-scene interactions from state-of-the-art video generation models, which have been trained on vast amounts of natural human movements and interactions, and use differentiable rendering to reconstruct human-scene interactions. \model can synthesize realistic human motions in both static scenes and environments with dynamic objects, without requiring any ground-truth motion data. We evaluate \model on a curated dataset of different types of various indoor and outdoor scenes with different interaction prompts, demonstrating its ability to generate diverse and contextually appropriate human-scene interactions.
Project page: \url{https://awfuact.github.io/zerohsi}.
\vspace{-12pt}

\end{abstract}

\section{Introduction}

Generating realistic human motions that interact with 3D environments is fundamental to computer graphics, VR/AR, embodied AI, and robotics. Humans constantly engage with their surroundings through both static interactions—sitting on chairs, lying on sofas, leaning against ladders—and dynamic interactions, such as watering plants, playing musical instruments, or manipulating objects. These interactions are remarkably diverse and pervasive in our daily lives, encompassing countless objects and countless ways of interacting with them. Despite significant advances in motion synthesis, realistically simulating this wide spectrum of human-scene interactions remains a fundamental challenge.

Prior work in human-scene interaction synthesis primarily follows two directions. The first focuses on interactions with static 3D scenes~\cite{hassan2021stochastic}, with recent advances driven by motion diffusion models~\cite{jiang2024scaling,jiang2024autonomous} trained on paired scenes and motion capture data. While these models can generate realistic motions for common activities like navigation and sitting, they struggle to generalize even within action categories. The second direction explores manipulation of dynamic objects~\cite{li2023object,li2025controllable,peng2023hoi}, showing success in generalizing within object categories but failing to handle significant geometric variations. Both approaches share a fundamental limitation: \textbf{they cannot synthesize interactions in unseen environments} (such as in-the-wild scenes or reconstructed scenes), as they rely on paired 3D scene and motion capture data, which are unavailable for unseen scenarios. This motivates us to address the problem of \emph{zero-shot human-scene interaction}, synthesizing HSIs in various scenes without training, and eliminating the need for 3D HSI data.

Our key insight is to distill human-scene interactions from state-of-the-art video generation models. These models have been trained on vast amounts of video data, capturing a wide range of natural human movements and interactions in diverse environments. This allows us to generate contextually appropriate human motions for various 3D scenes, whether synthesized or reconstructed. For example, given a reconstructed 3D garden scene, our method can generate natural motions of a person watering plants in the garden (Figure~\ref{fig:teaser}).

Our approach, \model, enables zero-shot 4D human-scene interaction synthesis by integrating video generation and neural human rendering. In a nutshell, \model takes a 3D scene as input, initializes an animatable human avatar in the scene, generates a video of the human interacting with the scene, and then extracts the interaction motion via differentiable neural rendering. \model can synthesize appropriate human motions in both static scenes and environments containing dynamic objects, without requiring any ground truth motion data.  Our design leverages pretrained video generation models, maintaining compatibility with various generation approaches and allowing seamless integration of future advancements in video generation. Our primary contributions can be summarized as follows:
\begin{itemize}
    \item We introduce the novel task of zero-shot HSI motion generation, addressing the fundamental limitation of requiring paired motion-scene training data.
    \item We propose \model, which integrates video generation with differentiable human rendering to tackle the challenging task, supporting both static scenes and environments with dynamic objects.
    \item We curate a dataset of various indoor and outdoor scenes with different interaction prompts and plausible initial human poses to evaluate zero-shot HSI. We demonstrate that \model can generate diverse and contextually appropriate human-scene interactions across these environments.
\end{itemize}

\vspace{-2pt}
\section{Related Work}
\vspace{-2pt}

\textbf{Text-Guided Motion Generation.} The availability of large-scale motion capture datasets like AMASS~\cite{mahmood2019amass}, enhanced with action labels and language descriptions through BABEL~\cite{punnakkal2021babel} and HumanML3D~\cite{guo2022generating}, has enabled significant advances in language-guided motion synthesis~\cite{guo2022generating,petrovich2022temos,tevet2022motionclip}. Early approaches demonstrated success using VAE architectures~\cite{guo2022tm2t,petrovich2021action}. More recently, diffusion models have emerged as a powerful framework for motion generation~\cite{chen2023executing,barquero2023belfusion,huang2023diffusion,raab2024single,shafir2024human,yuan2023physdiff,zhang2024tedi,tseng2023edge,li2023ego,li2023object,shi2023controllable}, leading to various text-conditioned approaches~\cite{tevet2023human,dabral2023mofusion,zhang2024motiondiffuse,karunratanakul2023gmd}. In contrast to these methods that focus on generating isolated human motions, our work synthesizes contextually appropriate human-scene interactions.

\myparagraph{Human-Scene Interaction Synthesis.} Research in human-scene interaction has progressed along multiple directions. With paired scene-motion datasets~\cite{hassan2019resolving,araujo2023circle,wang2022humanise,guzov2021human,zheng2022gimo} and object-motion data~\cite{zhang2022couch,hassan2021stochastic}, researchers have developed methods~\cite{wang2021synthesizing,wang2022towards,araujo2023circle,zhang2022couch,hassan2021stochastic,kulkarni2024nifty} for generating scene-aware human motions like sitting and reaching. In the domain of object manipulation, research has evolved from primarily focusing on hand motion synthesis~\cite{zhang2021manipnet,christen2022d,zheng2023cams} to incorporating full-body motions~\citep{taheri2022goal,wu2022saga,ghosh2023imos,li2024task,braun2024physically} with the fuel of full-body interaction datasets~\cite{taheri2020grab,fan2023arctic}. Building on paired human-object motion data~\cite{bhatnagar2022behave,wan2022learn,li2023object}, recent methods focus on predicting interactions from past states~\cite{wan2022learn,xu2023interdiff} or object motion sequences~\cite{li2023object}. While these approaches have shown promising results, they typically require paired motion-scene data. In contrast, we focus on zero-shot interaction synthesis, eliminating the need for scene-specific motion capture data while supporting diverse interaction in both static and dynamic environments. Besides interaction motion synthesis, prior work also explored zero-shot generation of static interactions at a single time step~\citep{kim2025beyond,li2024genzi}. These methods are complementary to our work as their output static poses can be used as the input initial poses in our approach.

Another line of work leverages reinforcement learning~\cite{lee2023locomotion,xiao2024unified} to train scene-aware policies for navigation and interaction in static 3D scenes, as well as object manipulation such as lifting and moving~\cite{hassan2023synthesizing,xie2023hierarchical,merel2020catch}. These approaches can be trained with relatively small amounts of motion data. However, they are limited to specific interaction types and restricted in generalization to diverse scenes and objects.

\myparagraph{Video Generation.} Recent advances in diffusion models have revolutionized video generation~\citep{hong2022cogvideo,singer2022make, ho2022imagen, blattmann2023align, girdhar2023emu, blattmann2023stable, bar2024lumiere, brooks2024video}, enabling high-quality generation of human actions~\citep{guo2023animatediff} and scene dynamics~\citep{brooks2024video}. Video generation models can now take both text and image as input conditions~\citep{yang2024cogvideox,kling}, where text descriptions guide the overall motion and actions while image conditions provide scene context and geometry, controlling fine-grained interaction details. These models have demonstrated remarkable capabilities in synthesizing temporally coherent videos, and recent efforts have extended video generation models to have increased controllability in terms of motion~\citep{wang2024motionctrl,guo2023animatediff} and camera viewpoints~\citep{he2024cameractrl}. Our work bridges the gap between generating realistic 2D videos and synthesizing 3D human motions. By integrating video generation as a drop-in module, our approach directly benefits from the rapid progress in video generation quality and controllability.

\myparagraph{Neural Rendering for Humans.} Neural rendering techniques have significantly advanced the synthesis of realistic human appearances~\citep{wang2022arah,liu2024animatable,lei2024gart,chen2023uv,kwon2024deliffas,jiang2022neuman,peng2021animatable,kocabas2024hugs,peng2021neural}. Building on the success of neural radiance fields (NeRF)~\cite{mildenhall2021nerf}, NeuralActor~\cite{liu2021neural} uses texture maps defined on the SMPL model~\cite{loper2015smpl} to guide the learning of deformable radiance fields. NeuralBody~\cite{peng2021neural} utilizes structured latent codes linked to SMPL~\cite{loper2015smpl} for novel view synthesis from sparse multi-view videos. AnimatableNeRF~\cite{peng2021animatable} introduces a neural blend weight field and achieves superior novel view and novel pose synthesis results. Additionally, efficiencies have been significantly improved for rendering avatars~\cite{jiang2023instantavatar} through multiresolution hash encoding representation~\cite{muller2022instant}. Recent studies have explored 3D Gaussian splatting~\cite{kerbl20233d}, an explicit and efficient representation for modeling animatable humans~\cite{li2024animatable,jiang2023instantavatar,yu2023monohuman,geng2023learning,hu2024gauhuman,zielonka25dega}. Animatable Gaussians~\cite{li2024animatable} leverages powerful 2D StyleGAN-based CNNs and 3D Gaussian splatting to create high-fidelity avatars from multi-view RGB videos. In our work, we leverage this differentiable and controllable representation to optimize SMPL parameters using an image-matching loss, bridging the gap between 2D video generation and 3D human motion. 

\begin{figure*}[t!]
    \centering
    \includegraphics[width=1\textwidth]{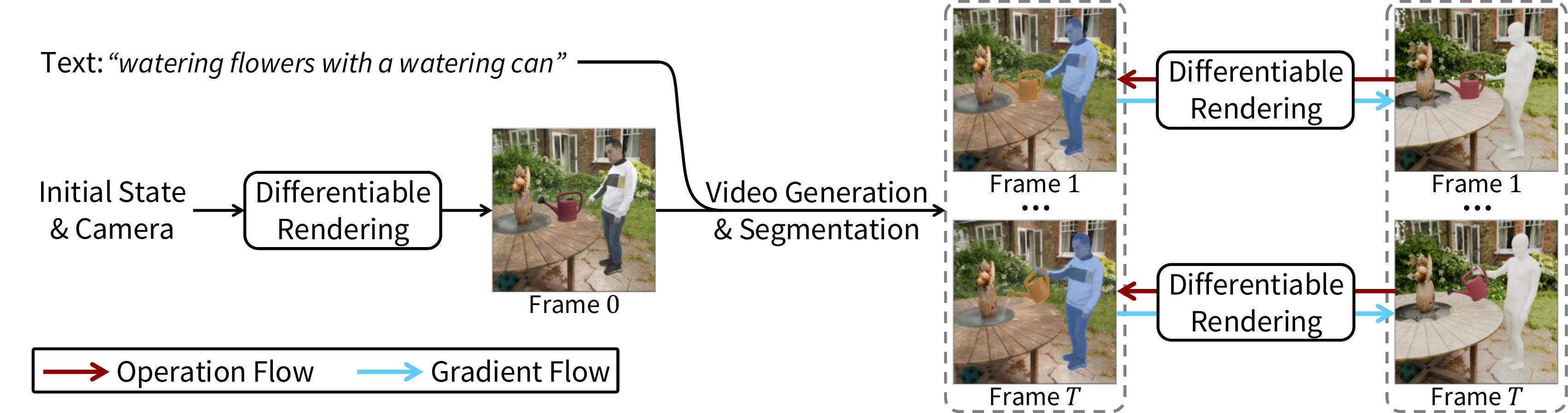}
    \vspace{-0.6cm}
    \caption{\textbf{Overview of \model.} Our approach begins with HSI video generation conditioned on the rendered initial state and text prompt. Through differentiable neural rendering, we optimize per-frame camera pose, human pose parameters, and object 6D pose by minimizing the discrepancy between the rendered and generated reference videos.}
    \label{fig:method}
    \vspace{-12pt}
\end{figure*}

\vspace{-2pt}
\section{\model}\label{sec:method}
\vspace{-2pt}

\textbf{Formulation.} Our goal is to generate a plausible 3D human-scene interaction motion sequence $\tau = \{(\mathcal{M}_t, \mathbf{P}_t)\}_{t=1}^T$, conditioned on a 3D scene $\mathcal{S}$, an interactable dynamic object $\mathcal{O}$, a text prompt describing the interaction $c$, and the initial states of human and objects $\tau_0$, where $\mathcal{M}_t$ represents the human pose in frame $t$, $\mathbf{P}_t \in \mathbb{R}^6$ represents the object's 6D pose in frame $t$, and $T$ represents the length of the sequence. We use SMPL-X~\cite{pavlakos2019expressive}, a widely used parameterized human model, to represent the human pose. In each frame, $\mathcal{M}_t$ consists of root translation $\mathbf{r}_t$, global orientation $\boldsymbol{\phi}_t$, and body pose $\mathbf{\Theta}_t$.

\noindent\textbf{Overview.} We present an illustration of our zero-shot human-scene interaction motion generation method in~\cref{fig:method}. To address the diversity and generalizability issues encountered by most existing learning-based methods, we first generate HSI video using off-the-shelf video generation models~\cite{kling} (\cref{sec:hsi-video-generation}). We then reconstruct the generated 2D HSI video into 4D human-scene-interaction sequence (\cref{sec:camera-pose-estimation,sec:hsi-optimization}). In addition, we incorporate a refinement process using human pose priors to improve the results (\cref{sec:refinement}).

The main challenge lies in converting the generated 2D HSI video into 4D interaction motion. While existing methods~\cite{cai2024smpler,shin2024wham,yin2024whac} can estimate plausible human motions, they have two key limitations for human-scene interaction scenarios. First, they struggle to estimate precise root translations, which often leads to penetration artifacts between the human and 3D scenes. Second, these methods do not address the reconstruction of object motions that are crucial for interaction with dynamic objects. To reconstruct plausible 4D HSI from a single video, we propose an optimization-based method using differentiable rendering. Specifically, we represent scene and object with 3D Gaussians~\cite{kerbl20233d}, and leverage a Gaussian Avatar model~\cite{li2024animatable} to map human poses to a set of Gaussian particles (\cref{sec:preliminaries}). We denote the Gaussian scene, object and human as $\mathcal{G}_\mathcal{S}$, $\mathcal{G}_\mathcal{O}$, and $\mathcal{G}_\mathcal{H}$, respectively. We then optimize the camera pose, human pose, and object's 6D pose through rendering loss. Additionally, we incorporate techniques to handle low-quality generated videos, ensuring robust performance with imperfect input footage.

\myparagraph{Discussion.} We justify the advantages of using off-the-shelf video generation model. Our method remains agnostic to video generation approaches, enabling compatibility with diverse models and easy integration of future advancements. By focusing on 2D-to-4D HSI lifting rather than being tied to a specific video generation model, we maintain flexibility to leverage ongoing research. We demonstrate this adaptability across different models in \cref{sec:other-model}.

\subsection{Preliminaries}\label{sec:preliminaries}

\textbf{3D Gaussian Splatting.} 3DGS~\cite{kerbl20233d} is an explicit 3D representation comprising a set of Gaussian particles, each characterized by its position $\boldsymbol{\mu}$ and covariance matrix $\mathbf{\Sigma}$:
\begin{equation}
    G(\mathbf{x})=e^{-\frac{1}{2}(\mathbf{x}-\boldsymbol{\mu})^\top\mathbf{\Sigma}^{-1}(\mathbf{x}-\boldsymbol{\mu})}.
\end{equation}
Each particle is parameterized by its position $\boldsymbol{\mu}$, opacity $\alpha$, rotation $\mathbf{r}$, scale $\mathbf{s}$, and color $\mathbf{c}$. The covariance matrix $\mathbf{\Sigma}$ is computed by $\mathbf{\Sigma}=\mathbf{R}\mathbf{S}\mathbf{S}^\top\mathbf{R}^\top$, where $\mathbf{R}$ is the rotation matrix constructed from $\mathbf{r}$, and $\mathbf{S}=\text{diag}([s_x,s_y,s_z])$ is the scaling matrix. The 3D Gaussians are splatted onto 2D plane during rendering, and the color is calculated by alpha-blending of $N$ ordered particles overlapping the pixel,
\begin{equation}
     \mathbf{C}=\sum_{i=1}^N\alpha_i\prod_{j=1}^{i-1}(1-\alpha_j)\mathbf{c}_i.\label{eq:gaussian-rasterization}
\end{equation}
We use $\mathcal{R}$ to represent this rendering process of 3DGS.

\myparagraph{Animatable Gaussians.} Animatable Gaussians~\cite{li2024animatable} is a neural human rendering method that maps SMPL-X parameters~\cite{pavlakos2019expressive} to an animatable avatar represented by 3DGS~\cite{kerbl20233d}. The appearance of the posed avatar can be differentiably rendered to an image $\mathbf{I}$ as in~\cref{eq:gaussian-rasterization}, with a given camera view $\mathbf{T}\in\mathbb{R}^{4\times4}$. Given the driving SMPL-X pose $\mathbf{\Theta}$, Animatable Gaussians first deforms the character-specific template using a linear blend skinning (LBS) function. It then predicts pose-dependent Gaussian maps conditioned on the driving pose $\mathbf{\Theta}$ and the camera view $\mathbf{T}$:
\begin{align}
    \{\Delta\boldsymbol{\mu}_p\}_{p=1}^{P}&\leftarrow\mathcal{F}_{position}(\mathbf{\Theta}), \\
    \{\alpha_p,\mathbf{s}_p,\mathbf{r}_p\}_{p=1}^{P}&\leftarrow\mathcal{F}_{other}(\mathbf{\Theta}), \\
    \{\mathbf{c}_p\}_{p=1}^{P}&\leftarrow\mathcal{F}_{color}(\mathbf{\Theta}, \mathcal{V}(\mathbf{T})), \label{eq:color-net}
\end{align}
where $\Delta\boldsymbol{\mu}_p$ is the offset position relative to the deformed template, and $\mathcal{V}$ is the view direction feature extractor. The root translation $\mathbf{r}$ and the global orientation $\boldsymbol{\phi}$ are finally applied to $\{\boldsymbol{\mu}_p,\mathbf{s}_p\}_{p=1}^{P}$ to achieve global transformation. We denote the mapping from human pose and camera view to Gaussian particles proposed by Animatable Gaussians as
\begin{equation}
     \mathcal{G}_\mathcal{H}\leftarrow\mathcal{A}(\mathbf{r},\boldsymbol{\phi},\mathbf{\Theta};\mathbf{T}).\label{eq:animatable-gaussian}
\end{equation}

\subsection{HSI Video Generation and Processing}\label{sec:hsi-video-generation}

\begin{figure}[t!]
    \centering
    \includegraphics[width=0.47\textwidth]{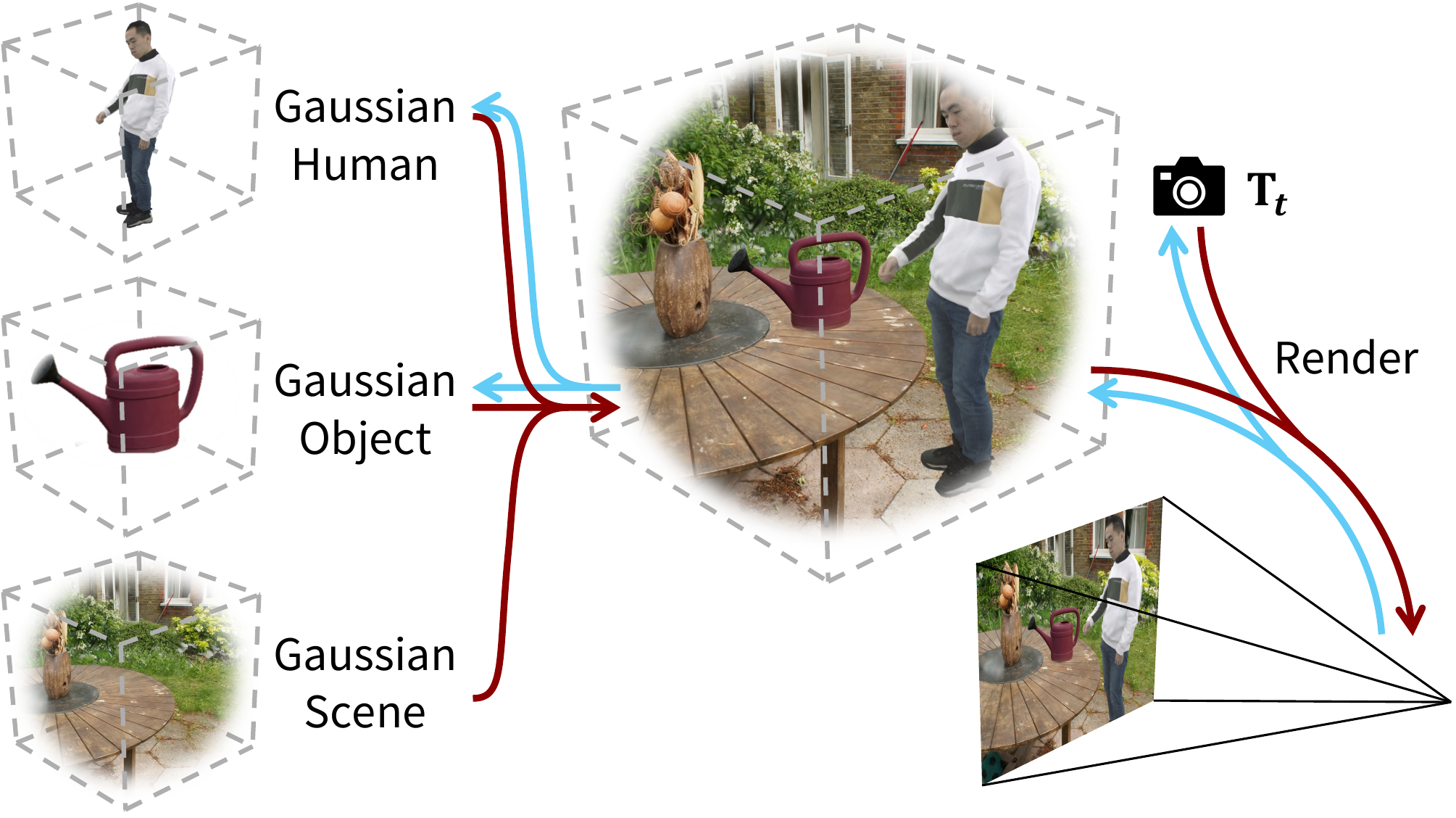}
    \vspace{-0.2cm}
    \caption{\textbf{Illustration of the differentiable rendering process.} The parameterized Gaussian human, transformed Gaussian object, and static Gaussian scene are concatenated and rendered through Gaussian rasterization.}
    \label{fig:render}
    \vspace{-12pt}
\end{figure}

Given the initial human pose $\mathcal{M}_0=(\mathbf{r}_0,\boldsymbol{\phi}_0,\mathbf{\Theta}_0)$ and an additional camera pose input $\mathbf{T}_0$, we first map them to the initial human Gaussians $\mathcal{G}_\mathcal{H}^0$ through~\cref{eq:animatable-gaussian}. The initial object 6D pose $\mathbf{P}_0$ is then applied to the Gaussian object $\mathcal{G}_{\mathcal{O}}$, denoted as $\mathcal{G}_{\mathcal{O}}^0=\mathcal{G}_{\mathcal{O}}(\mathbf{P}_0)$. As depicted in \cref{fig:render}, the resulting Gaussians are concatenated with the Gaussian scene $\mathcal{G}_{\mathcal{S}}$ and rendered into the initial HSI image $\mathbf{I}_0$ through~\cref{eq:gaussian-rasterization} under camera view $\mathbf{T}_0$.

The HSI video $\{\mathbf{I}_t\}_{t=0}^T$ is generated through the KLING image-to-video model~\cite{kling}, conditioned on the rendered initial frame $\mathbf{I}_0$ and text prompt $c$. We employ Segment Anything Model 2 (SAM2)~\cite{ravi2024sam} to segment dynamic foreground (human and object) and static background within the generated video for subsequent uses. The prompt points for SAM2 in the initial frame are automatically proposed through farthest point sampling on non-occluded regions of both human and object. We denote the segmentation masks for human and object as $\{\mathbf{M}_{\mathcal{H}}^t\}_{t=0}^T$ and $\{\mathbf{M}_{\mathcal{O}}^t\}_{t=0}^T$, respectively.

\myparagraph{Handling Occlusions.} We address the challenge of under-constrained pose parameters caused by body part occlusions by minimizing occlusions in the generated video. Since camera movement remains minimal over short durations (e.g., 5 seconds), we simply select an initial camera view $\mathbf{T}_0$ from a pre-defined camera array that maximizes visibility of both the human and dynamic object in the first frame. Details of the camera view selection process are in \cref{sec:camera-view-selection}.

\subsection{Camera Pose Estimation}\label{sec:camera-pose-estimation}

We sequentially estimate the relative camera transformation between nearby frames for camera pose estimation. In frame $t$, we apply a learnable transformation $\mathbf{T}\in\mathbb{R}^{4\times4}$ to the Gaussian scene $\mathcal{G}_{\mathcal{S}}$, denoted as $\mathcal{G}_{\mathcal{S}}(\mathbf{T})$. We render image with the estimated camera pose $\mathbf{T}_{t-1}$, and optimize the relative transformation $\mathbf{T}$ with the photometric loss between frame $t$, focusing on the static background region:
\begin{equation}
    \mathbf{T}_*=\arg\min_{\mathbf{T}}\mathcal{L}_2\Big(\mathcal{R}\big(\mathcal{G}_{\mathcal{S}}(\mathbf{T});\mathbf{T}_{t-1}\big)\odot\mathbf{M}_t, \mathbf{I}_t\odot\mathbf{M}_t\Big),
\end{equation} 
where $\mathbf{M}_t$ represents the mask of static background. The camera pose for frame $t$ is calculated by $\mathbf{T}_t=\mathbf{T}_*^{-1}\mathbf{T}_{t-1}$.

\myparagraph{Handling Incorrect Contents.} Video generation models produce incorrect contents in regions initially occluded by humans or objects that later become visible (\cref{sec:incorrect-contents}). Instead of using only the current frame's static background mask, we address this by aggregating dynamic foreground masks from frames 0 to $t$ to eliminate incorrect contents:
\begin{equation}
    \mathbf{M}_t=\mathbf{1}-\bigcup_{i=0}^{t}\big(\mathbf{M}_{\mathcal{H}}^i\cup\mathbf{M}_{\mathcal{O}}^i\big).
\end{equation}

\subsection{HSI Optimization}\label{sec:hsi-optimization}

The optimization of 4D human-scene interactions is carried out sequentially on a frame-by-frame basis. For object 6D pose estimation, we also optimize the relative transformation $\mathbf{P}\in\mathbb{R}^6$ between nearby frames. We optimize the human pose with a different strategy, as optimizing relative transformations leads to rapidly increased cumulative errors, particularly when the generated video exhibits quality issues such as body part disappearance (\cref{sec:body-part-disappearance}). We directly optimize the human pose parameters $\mathcal{M}_t=(\mathbf{r}_t,\boldsymbol{\phi}_t,\mathbf{\Theta}_t)$, where the root translation $\mathbf{r}_t$ and global orientation $\boldsymbol{\phi}_t$ are initialized using their respective values $\mathbf{r}_{t-1}$ and $\boldsymbol{\phi}_{t-1}$ from the previous frame. Additionally, the body poses $\{\mathbf{\Theta}_t\}_{t=1}^T$ are initialized using independent frame-wise estimates from a pose estimation model~\cite{cai2024smpler}.

Similar to~\cref{sec:hsi-video-generation}, we render the Gaussians human, object, and scene at frame $t$ through
\begin{align}
    \hat{\mathbf{I}}_t&=\mathcal{R}(\mathcal{G}_{\mathcal{H}}^t,\mathcal{G}_{\mathcal{O}}^t,\mathcal{G}_{\mathcal{S}};\mathbf{T}_t)\nonumber\\
    &=\mathcal{R}\big(\mathcal{A}(\mathbf{r}_t,\boldsymbol{\phi}_t,\mathbf{\Theta}_t;\mathbf{T}_t), \mathcal{G}_{\mathcal{O}}(\mathbf{P}_t), \mathcal{G}_{\mathcal{S}};\mathbf{T}_t\big),
\end{align}
where $\mathbf{P}_t$ denotes the object's 6D pose in frame $t$, computed as the composition of the relative transformation $\mathbf{P}$ and the previously optimized object pose $\mathbf{P}_{t-1}$.

Using the generated video as the reference, we optimize the human pose parameters $\mathcal{M}_t$ with the photometric loss in each frame:
\begin{equation}
    \mathcal{L}_{\text{rgb}}=(1-\lambda)\mathcal{L}_1(\hat{\mathbf{I}}_t,\mathbf{I}_t)+\lambda\mathcal{L}_{\text{D-SSIM}}(\hat{\mathbf{I}}_t,\mathbf{I}_t).\label{eq:photometric-loss}
\end{equation}

We introduce two additional loss terms to enhance the accuracy of object 6D pose optimization. For frame $t$, we compute the center point position $C_{\mathcal{O}}^t\in\mathbb{R}^2$ of the object's segmentation mask $\mathbf{M}_{\mathcal{O}}^t$ and the center point position of the rendered object region $\hat{C}_{\mathcal{O}}^t$. These positions are normalized to the range [0,1], and the object center point position loss is defined as
\begin{equation}
    \mathcal{L}_{\text{center}}=\mathcal{L}_2(\hat{C}_{\mathcal{O}}^t,C_{\mathcal{O}}^t).
\end{equation}
A depth regularization term is incorporated based on the assumption that the object's depth remains relatively constant throughout a short time window (e.g., 5 seconds), and thus, the object depth within the time window should be close to the depth value at the first frame. The average object depth in the first frame is computed as
\begin{equation}
    D_{\mathcal{O}}^0=\frac{\texttt{sum}(\mathbf{D}_0\odot\mathbf{M}_{\mathcal{O}}^0)}{\texttt{sum}(\mathbf{M}_{\mathcal{O}}^0)}\in\mathbb{R}, 
\end{equation}
where $\mathbf{D}_0$ represents the rendered depth map in frame 0. The average depth of the rendered object region $\hat{D}_{\mathcal{O}}^t$ is calculated similarly, and we define the depth regularization loss as
\begin{equation}
    \mathcal{L}_{\text{depth}}=\mathcal{L}_2(\hat{D}_{\mathcal{O}}^t,D_{\mathcal{O}}^0).
\end{equation}

The object's relative transformation $\mathbf{P}$ is optimized using the composite loss function:
\begin{equation}
    \mathcal{L}=\mathcal{L}_{\text{rgb}}+\lambda_{\text{center}}\mathcal{L}_{\text{center}}+\lambda_{\text{depth}}\mathcal{L}_{\text{depth}}.
\end{equation}

\myparagraph{Handling Appearance Change.} Video generation models do not guarantee a consistent human appearance (\cref{sec:appearance-change}), which can potentially compromise human pose optimization. Given that human appearance exhibits continuous changes in the generated video, we fine-tune the color net defined in~\cref{eq:color-net} during optimization using $\mathcal{L}_{\text{rgb}}$.

\subsection{Refinement}\label{sec:refinement}

Limited supervision from single-view video poses challenges for natural human motion reconstruction. We refine our results in the latent space of VPoser~\cite{pavlakos2019expressive}, a variational human pose prior trained on the AMASS dataset~\cite{mahmood2019amass}, which preserves robust human pose priors. For each frame, the reference joint positions $\hat{J}_t$ are computed by passing the raw human pose parameters obtained through~\cref{sec:hsi-optimization} to the SMPL-X layer~\cite{pavlakos2019expressive}. We then optimize the root translation $\mathbf{r}_t$, global orientation $\boldsymbol{\phi}_t$, and the body pose latent $\mathbf{z}_t\in\mathbb{R}^{32}$ with the fitting loss:
\begin{equation}
    \mathcal{L}_{\text{fit}}^t=\mathcal{L}_2\Big(\hat{J}_t, J_t\big(\mathbf{r}_t,\boldsymbol{\phi}_t,\mathcal{D}(\mathbf{z}_t)\big)\Big),
\end{equation}
where $\mathcal{D}$ denotes the VPoser decoder. Following prior works~\cite{huang2023diffusion,xu2023interdiff,xu2024interdreamer}, additional physics losses are incorporated to enhance physical plausibility, yielding the composite loss:
\begin{equation}
    \mathcal{L}=\frac{1}{T}\sum_{t=0}^T\mathcal{L}_{\text{fit}}^t+\lambda_{\text{physics}}\mathcal{L}_{\text{physics}}.
\end{equation}
Detailed formulation of $\mathcal{L}_{\text{physics}}$ is in \cref{sec:loss}.

\begin{figure*}[t!]
    \centering
    \includegraphics[width=1\textwidth]{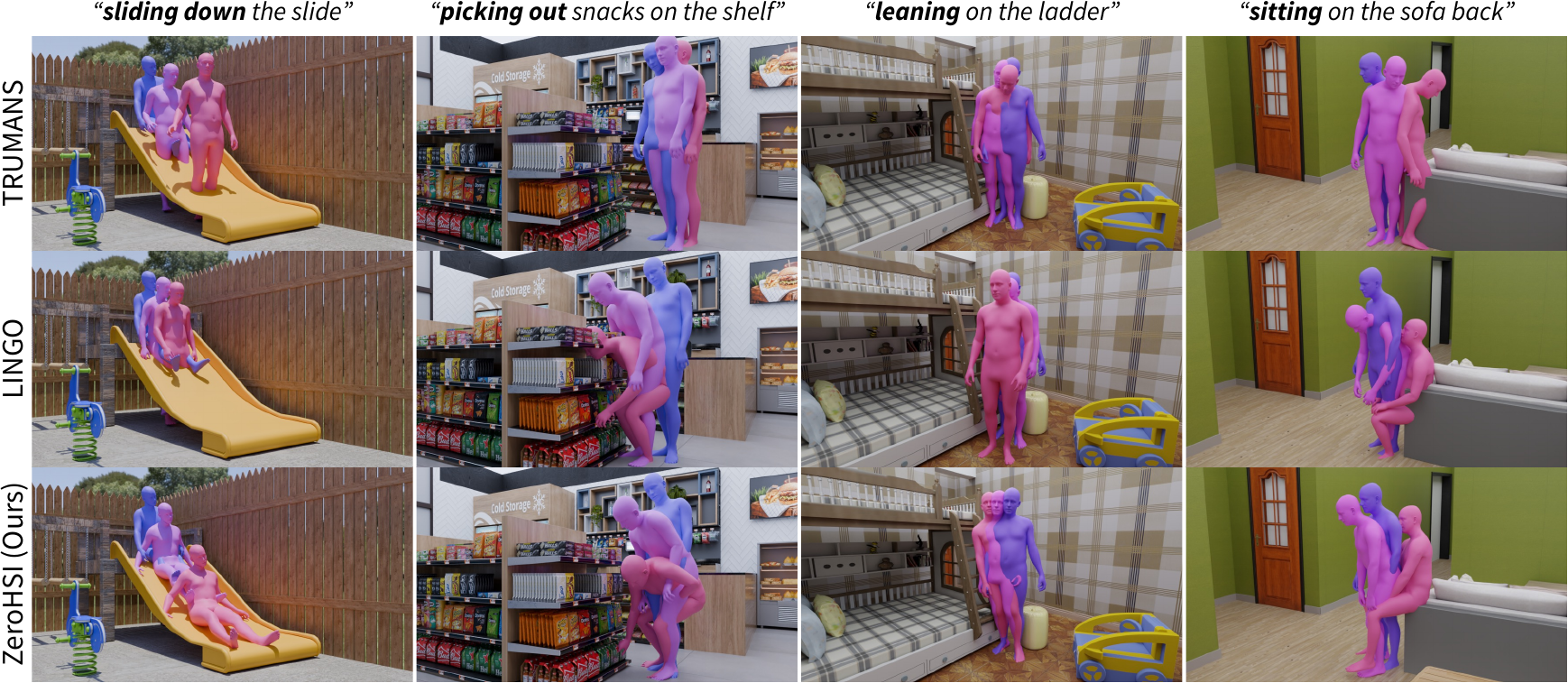}
    \vspace{-0.6cm}
    \caption{\textbf{Qualitative comparison of interactions with static scenes on \dataset.} \model generates 4D HSIs that are more realistic and better aligned with text prompts, demonstrating generalizability across diverse scenes and interaction types compared to baselines.}
    \label{fig:results-static}
\end{figure*}

\begin{figure*}[t!]
    \centering
    \includegraphics[width=1\textwidth]{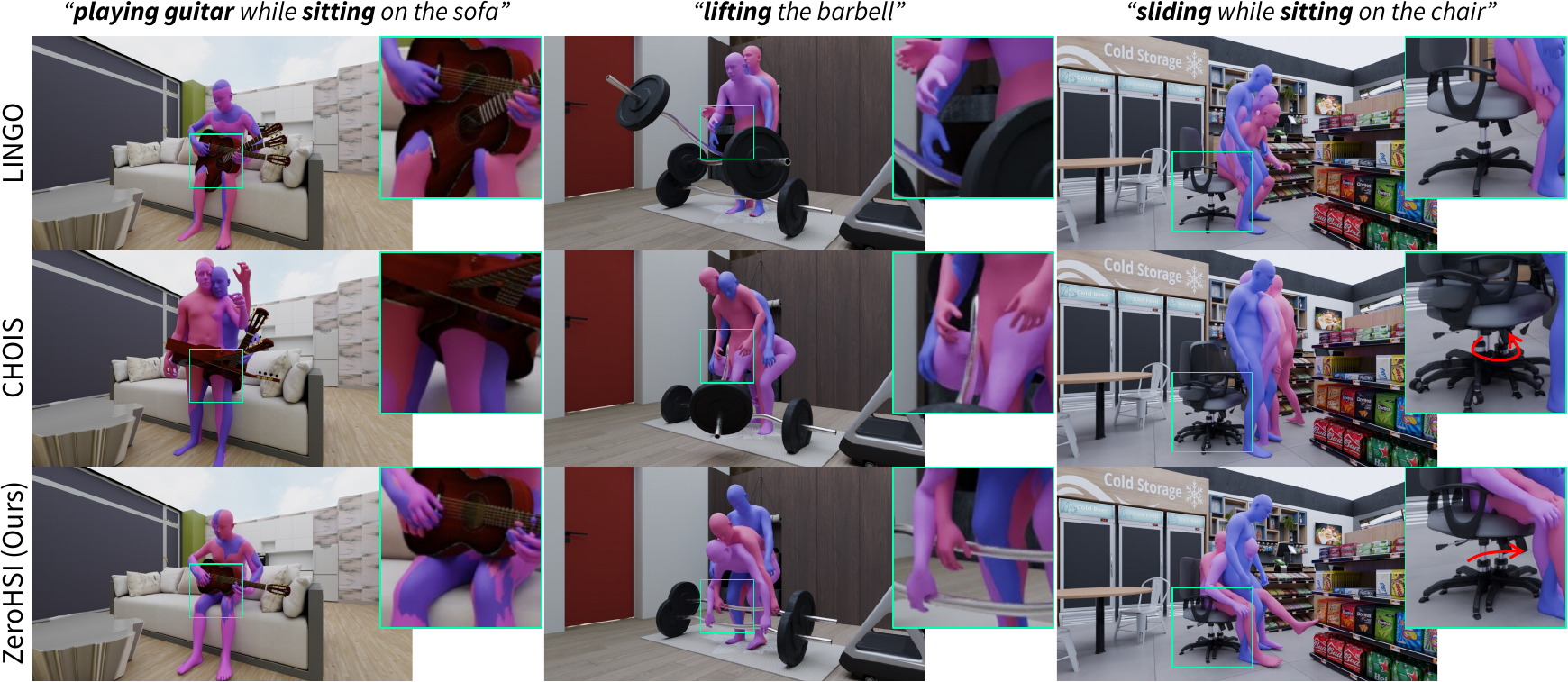}
    \vspace{-0.6cm}
    \caption{\textbf{Qualitative comparison of interactions with dynamic objects in scenes on \dataset.} Our method maintains proper object contact while minimizing penetration, successfully handling challenging interactions like sliding while seated on an office chair.}
    \label{fig:results-dynamic}
    \vspace{-12pt}
\end{figure*}

\begin{figure*}[ht!]
    \centering
    \includegraphics[width=1\textwidth]{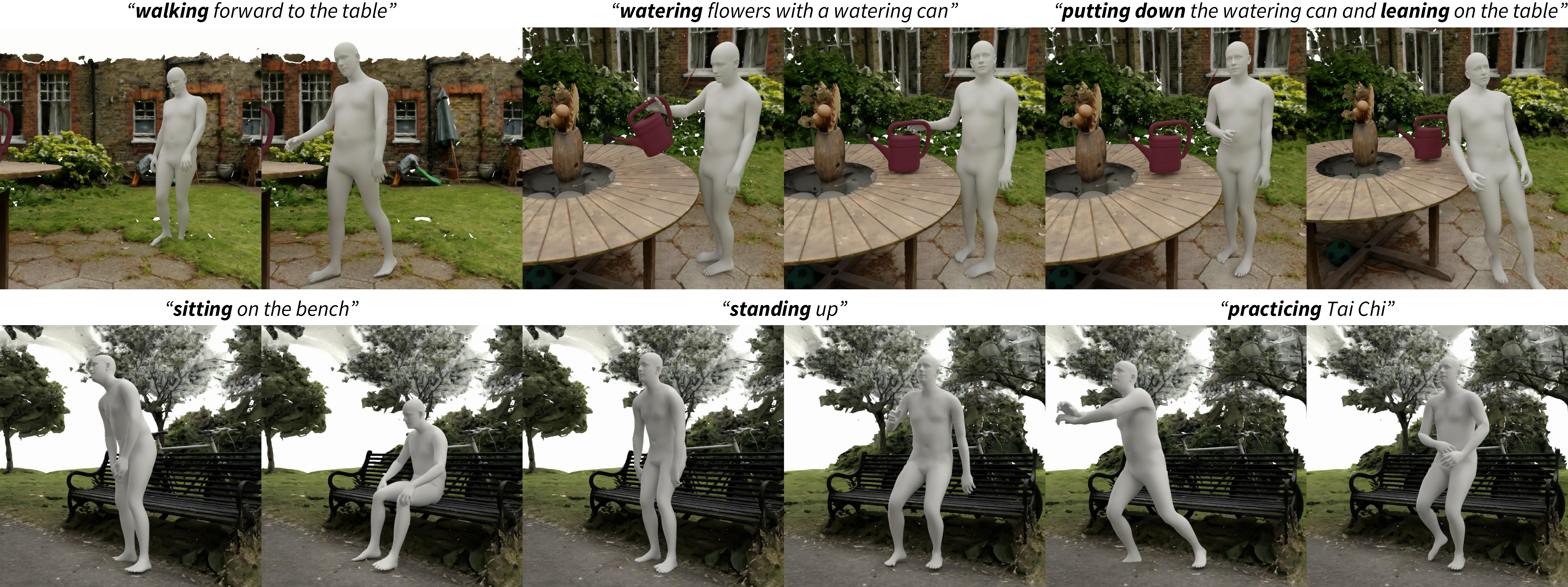}
    \vspace{-0.6cm}
    \caption{\textbf{Qualitative results of long-term interactions with reconstructed real scenes on \dataset.} \model generates long-term interaction sequences with multiple text prompts in reconstructed scenes from the Mip-NeRF 360 (Garden and Bicycle) dataset~\cite{barron2022mip}. 
    }
    \label{fig:long-short}
    \vspace{-12pt}
\end{figure*}

\section{Experiments}

\textbf{Experimental Settings.} We evaluate our HSI generation framework across two distinct settings: static scene interactions and dynamic object interactions. The \textit{\textbf{static setting}} focuses on human motion synthesis within fixed environments, where the scene geometry remains unchanged throughout the interaction. The \textit{\textbf{dynamic setting}} extends this to include movable objects, requiring the simultaneous generation of human motion and object pose sequences.

\noindent\textbf{Evaluation Dataset.} We introduce \dataset, an evaluation benchmark comprising various 3D environments sourced from existing datasets~\cite{jiang2024scaling}, public asset libraries, and real-world reconstructed scenes from Mip-NeRF 360~\cite{barron2022mip} and Tanks and Temples~\cite{knapitsch2017tanks}. The dataset encompasses 12 distinct 3D environments (7 indoor and 5 outdoor), spanning residential spaces (e.g., bedrooms and living rooms), recreational facilities (e.g., gyms and cafes), and outdoor venues (e.g., greenhouses and playgrounds). For evaluating the dynamic setting, we augment these environments with interactive object models from BlenderKit and 3dsky. Each scene is annotated with 1-3 natural language descriptions of human-scene interactions and corresponding initial positions. To assess generative diversity, we synthesize multiple HSI sequences per text-position pair, yielding 100 distinct evaluation instances. Details and statistics of our \dataset dataset are available in \cref{sec:dataset}.

\noindent\textbf{Baselines.} For the the static setting, we use TRUMANS~\cite{jiang2024scaling} and LINGO~\cite{jiang2024autonomous} as baseline methods. TRUMANS~\cite{jiang2024scaling} generates interactions with static scenes conditioned on a navigation trajectory. LINGO~\cite{jiang2024autonomous} synthesizes scene-aware human motions autonomously based on text instructions and goal locations inside the scene.

In the dynamic setting, we compare our method with LINGO~\cite{jiang2024autonomous} and CHOIS~\cite{li2025controllable}. LINGO~\cite{jiang2024autonomous} generates grasp/put-down actions by designating goal locations for hand-object attachment and release. CHOIS~\cite{li2025controllable}  synthesizes human-object interaction guided by language and sparse object waypoints.

\myparagraph{Evaluation Metrics.} Since we target a zero-shot generation task that does not have ground-truth, we evaluate our approach through rendered videos and quantitative metrics assessing three key aspects: semantic alignment, motion diversity, and physical plausibility. We render the synthesized interactions by applying the generated SMPL-X parameters and 6D object pose sequences to Gaussian avatar and object, visualizing the results via Gaussian rasterization~\cite{kerbl20233d}.

\noindent
\underline{Semantic alignment:} We compute: (i) CLIP score~\cite{radford2021learning} between input text prompts and rendered frames to assess text-motion correspondence, and (ii) frame-wise CLIP consistency to measure temporal coherence through cosine similarity of adjacent frame embeddings.

\noindent
\underline{Motion diversity:} We generate five HSI sequences per evaluation instance with identical inputs. We compute the mean per-joint Euclidean distance between each pair of generated sequences, with higher values indicating greater diversity in the synthesized motions.

\noindent
\underline{Physical plausibility:} For both static and dynamic settings, we measure scene penetration following~\cite{zhao2023synthesizing}: (i) Pene$_{\%scene}$: percentage of body vertices penetrating the scene, (ii) Pene$_{mean}$: average penetration depth, and (iii) Pene$_{max}$: maximum penetration depth. For static settings, we additionally evaluate foot sliding score (FS) adapted from NeMF~\cite{he2022nemf}. In dynamic settings, we assess object interactions using CHOIS~\cite{li2025controllable} metrics: hand-object contact ratio (Cont.) and average object penetration depth (Pene$_{obj}$). All penetration metrics are computed using pre-calculated Signed Distance Fields (SDFs) for scenes and objects.

\begin{table*}[ht!]
    \centering
    \small
    \begin{tabular}{lccccccc}
        \toprule
        Method & CLIP Score$\uparrow$ & CLIP Consistency$\uparrow$ & Diversity$\uparrow$ & Pene$_{\%scene}$$\downarrow$  & Pene$_{mean}$$\downarrow$  & Pene$_{max}$$\downarrow$  & FS$\downarrow$ \\
        \midrule
        TRUMANS~\cite{jiang2024scaling} & 22.36 & \underline{0.9934} & 0.1527 & \underline{0.046} & \underline{0.240} & \underline{1.892} & 0.196 \\
        LINGO~\cite{jiang2024autonomous} & \underline{22.61} & \textbf{0.9942} & \underline{0.1698} & 0.058 & 0.421 & 2.056 & \textbf{0.106} \\
        \model (ours) & \textbf{23.52} & 0.9928 & \textbf{0.1703} & \textbf{0.019} & \textbf{0.147} & \textbf{1.330} & \underline{0.158} \\
        \bottomrule
    \end{tabular}
    \vspace{-0.25cm}
    \caption{\textbf{Quantitative evaluation of interactions with static scenes.} \model achieves better semantic alignment with text inputs (higher CLIP score), motion diversity, and physical plausibility (lower scene penetration) compared to TRUMANS~\cite{jiang2024scaling} and LINGO~\cite{jiang2024autonomous}.}
    \label{tab:static-scene}
    \vspace{-2mm}
\end{table*}

\begin{table*}[ht!]
    \centering
    \small
    \setlength{\tabcolsep}{4pt}
    \begin{tabular}{lcccccccc}
        \toprule
        Method & CLIP Score$\uparrow$ & CLIP Consistency$\uparrow$ & Diversity$\uparrow$ & Pene$_{\%scene}$$\downarrow$  & Pene$_{mean}$$\downarrow$  & Pene$_{max}$$\downarrow$  & Cont.$\uparrow$ & Pene$_{obj}$$\downarrow$ \\
        \midrule
        CHOIS~\cite{li2025controllable} & 22.11 & 0.9871 & \textbf{0.3382} & 0.025 & 0.191 & 1.877 & \underline{0.687} & \underline{1.581} \\
        LINGO~\cite{jiang2024autonomous} & \underline{22.99} & \textbf{0.9965} & 0.0914 & 0.032 & \textbf{0.089} & \textbf{0.446} & 0.699 & 0.242 \\
        \model (ours) & \textbf{24.01} & \underline{0.9955} & \underline{0.1942} & \textbf{0.022} & \underline{0.109} & \underline{1.062} & \textbf{0.835} & \textbf{0.033} \\
        \bottomrule
    \end{tabular}
    \vspace{-0.25cm}
    \caption{\textbf{Quantitative evaluation of interactions with dynamic objects in scenes.} Our method outperforms baselines with stronger semantic alignment with text prompts and better dynamic interaction quality (higher contact ratio and lower object penetration).}
    \label{tab:dynamic-object}
    \vspace{-4mm}
\end{table*}

\begin{table}[t!]
    \centering
    \small
    \setlength{\tabcolsep}{7pt}
    \begin{tabular}{llcc}
        \toprule
        Setting & Method & Realism & Alignment \\
        \midrule
        \multirow{2}{*}{Static} & vs. TRUMANS~\cite{jiang2024scaling} & 85.0\% & 93.1\% \\
        & vs. LINGO~\cite{jiang2024autonomous} & 75.1\% & 84.0\% \\
        \midrule
        \multirow{2}{*}{Dynamic} & vs. CHOIS~\cite{li2025controllable} & 96.5\% & 99.0\% \\
        & vs. LINGO~\cite{jiang2024autonomous} & 86.9\% & 89.1\% \\
        \bottomrule
    \end{tabular}
    \vspace{-2.5mm}
    \caption{\textbf{Human study on generated 4D HSI motions.} In both static and dynamic scenarios, participants prefer our generated HSIs for their motion realism and semantic alignment by large margins.}
    \label{tab:human-study}
    \vspace{-5mm}
\end{table}

\vspace{-2pt}
\subsection{Comparisons}
\vspace{-2pt}

\textbf{Static Setting.} We show examples of zero-shot HSI generation in~\cref{fig:results-static}. TRUMANS~\cite{jiang2024scaling}, which only accepts scene conditions as input in our experiments, merely follows the trajectory without meaningful interactions when encountering novel scenes. While LINGO~\cite{jiang2024autonomous} successfully generates text-aligned HSIs for familiar text prompts, it fails to properly avoid collisions in unseen environments, as indicated in examples of picking snacks and sitting on sofa backs. In addition, LINGO also struggles to synthesize reasonable motions of unseen interaction types. In contrast, \model generates plausible 4D HSIs across diverse scenes, demonstrating its generalizability to various environments.

We show the quantitative metrics in \cref{tab:static-scene}. Aligned with visual observations, our \model outperforms TRUMANS~\cite{jiang2024scaling} and LINGO~\cite{jiang2024autonomous} across most evaluation metrics, achieving the highest CLIP score and diversity while significantly reducing scene penetration metrics. These results indicate that our approach generates more diverse and plausible motions with better semantic alignment.

\myparagraph{Dynamic Setting.} In terms of the dynamic setting, our method excels in interactions involving dynamic objects. As illustrated in \cref{fig:results-dynamic}, while LINGO~\cite{jiang2024autonomous} and CHOIS~\cite{li2025controllable} generate semantically relevant motions, they suffer from object penetration issues (playing guitar) and poor contact quality (lifting barbell). They also struggle with complex interactions requiring multiple body parts and global translation, such as sliding while seated. Our \model consistently handles these challenging settings, generating high-quality interactions with dynamic objects.

We show the quantitative results in~\cref{tab:dynamic-object}. \model achieves the highest contact ratio and lowest object penetration compared to LINGO~\cite{jiang2024autonomous} and CHOIS~\cite{li2025controllable}. This precise object interaction is further complemented by strong semantic alignment and reliable scene interaction, demonstrating our method's capability to generate natural and accurate interactions with dynamic objects within scenes while maintaining high motion quality.

\myparagraph{Human Perceptual Study.} We conduct a human perceptual study using the two-alternative forced choice (2AFC) method. We recruit 400 participants. Participants are given results generated by different methods using identical input and instructed to select the sample they perceive as more realistic and better aligned with the textual description. As reported in \cref{tab:human-study}, users consistently prefer our results over baselines by significant margins in both static and dynamic settings, aligning well with our quantitative metrics and qualitative examples.

\vspace{-4pt}
\subsection{Long-Term Interaction Synthesis}
\vspace{-4pt}

We show examples of long-term HSIs in reconstructed real scene generated by \model in \cref{fig:long-short}. The Garden example shows a character walks forward, waters flowers, puts down the watering can, and leans on the table. In the Bicycle example, the character sits on a bench, stands up, and practices Tai Chi. These examples showcase \model's capability to generate realistic interactions in real-world environments and its application for producing sequences of varying lengths. More details and visualizations are in \cref{sec:long-term}.

\vspace{-4pt}
\subsection{Ablation Study}
\vspace{-4pt}

To evaluate the effectiveness of our per-frame optimization approach, we conduct an ablation study in the static scenario. We replace our camera pose estimation and HSI optimization modules with state-of-the-art human motion reconstruction methods WHAM~\cite{shin2024wham} and WHAC~\cite{yin2024whac}. Additionally, we evaluate a variant of our model (``\model w/o OPT$_\text{body}$'') that removes body pose optimization and instead only optimizes root translation and global orientation for 30 iterations per frame, using poses directly estimated by SMPLer-X~\cite{cai2024smpler}. 

Quantitative results in~\cref{tab:ablation-study} show that our full model outperforms the ablated variants across most metrics, demonstrating that our per-frame optimization approach effectively reconstructs human-scene interactions and serves as the optimal solution for bridging 2D HSI videos and 4D HSIs. Qualitative examples are presented in \cref{sec:ablation-qualitative}.

\begin{table}[t!]
    \small
    \begin{center}
        \footnotesize{
            \setlength{\tabcolsep}{2.7pt}
            \begin{tabular}{lcccccc}
                \toprule
                Method & CS$\uparrow$ & CC$\uparrow$ & P$_{\%}$$\downarrow$  & P$_{mean}$$\downarrow$  & P$_{max}$$\downarrow$ & FS$\downarrow$ \\
                \midrule
                WHAM~\cite{shin2024wham} & 22.73 & 0.9890 & 0.048 & 0.505 & 4.917 & \textbf{0.054} \\
                WHAC~\cite{yin2024whac} & 22.81 & \underline{0.9903} & 0.168 & 15.30 & 99.68 & 0.985 \\
                \model w/o OPT$_\text{body}$ & \underline{23.39} & 0.9890 & \underline{0.025} & \underline{0.165} & \textbf{1.092} & 0.245 \\
                \model (ours) & \textbf{23.52} & \textbf{0.9928} & \textbf{0.019} & \textbf{0.147} & \underline{1.330} & \underline{0.158} \\
                \bottomrule
            \end{tabular}
        }
    \end{center}
    \vspace{-6mm}
    \caption{\textbf{Quantitative results of ablation study.} ``CS'' denotes CLIP Sore, ``CC'' denotes CLIP Consistency, ``P$_{\%}$'' denotes Pene$_{\%scene}$, ``P$_{mean}$'' denotes Pene$_{mean}$, ``P$_{max}$'' denotes Pene$_{max}$. Our full method achieves higher motion quality and physical plausibility compared to variants.}
    \label{tab:ablation-study}
    \vspace{-6mm}
\end{table}

\vspace{-4pt}
\section{Conclusion}
\vspace{-4pt}

We present \model, a zero-shot approach to 4D human-scene interaction generation that addresses the limitation of requiring paired motion-scene training data. Our method successfully distills HSIs from video generation models through neural human rendering to synthesize contextually appropriate interactions across diverse environments.

\noindent\textbf{Acknowledgments.} We thank Zimo He and Nan Jiang for experimental setup. The work was in part supported by ONR YIP N00014-24-1-2117 and ONR MURI N00014-22-1-2740. J. Li was in part supported by the Wu Tsai Human Performance Alliance at Stanford University.

\pagebreak

\clearpage

{
    \small
    \bibliographystyle{ieee_fullname}
    \bibliography{main}
}

\clearpage
\appendix
\renewcommand\thefigure{S\arabic{figure}}
\setcounter{figure}{0}
\renewcommand\thetable{S\arabic{table}}
\setcounter{table}{0}
\renewcommand\theequation{S\arabic{equation}}
\setcounter{equation}{0}
\pagenumbering{arabic}%
\renewcommand*{\thepage}{S\arabic{page}}
\setcounter{footnote}{0}
\setcounter{page}{1}

\section{Overview}

In this supplementary material, we provide additional details on the dataset (\cref{sec:dataset}), experiments (\cref{sec:exp}), generated videos (\cref{sec:low-quality-video}), and an overall algorithm (\cref{sec:alg}) of \model. We highly recommend viewing our \href{https://awfuact.github.io/zerohsi/}{\textit{project page}} for compelling demonstrations across diverse scenarios.

\section{\dataset Dataset}\label{sec:dataset}

In this section, we elaborate on the statistics of the \dataset dataset. We summarize these statistics in \cref{tab:dataset-statistics} and visualize our \dataset dataset in \cref{fig:dataset-visualization}.

\subsection{Scenes}

Our \dataset dataset consists of diverse scenes from TRUMANS dataset~\cite{jiang2024scaling}, public 3D assets libraries, and reconstructed real scenes from the Mip-NeRF 360 dataset~\cite{barron2022mip} and Tanks and Temples dataset~\cite{knapitsch2017tanks}, resulting in $7$ indoor scenes (\textit{Bedroom}, \textit{Living Room}, \textit{Gym}, \textit{Bar}, \textit{Greenhouse}, \textit{Store}, \textit{Room}) and $5$ outdoor scenes (\textit{Playground}, \textit{Cafe}, \textit{Garden}, \textit{Bicycle}, \textit{Truck}). 

Among our synthetic scenes, \textit{Playground} and \textit{Cafe} are manually composed using models from 3D asset libraries, while the remaining six scenes are sourced directly from asset libraries with adjustments to their scale and layout to facilitate interactions. For 3DGS~\cite{kerbl20233d} reconstruction, we generate $300$-$500$ cameras per scene and manually filter occluded and low-quality views. We then render RGBD images to obtain initial point clouds and perform the reconstruction using the official 3D Gaussian Splatting implementation~\cite{kerbl20233d}.

For real scenes, we reconstruct them using images and camera views from the official datasets. We scale and transform these scenes to align their ground level and match real-world sizes. For visualization, we further extract scene meshes using SuGaR~\cite{guedon2024sugar}.

\subsection{Dynamic Objects}

\dataset includes $7$ types of dynamic objects (\textit{Guitar}, \textit{Barbell}, \textit{Watering Can}, \textit{Office Chair}, \textit{Shopping Cart}, \textit{Vase}, \textit{Mower}), and all of them are rigid. We obtain these objects from public 3D Assets libraries. Similar to the reconstruction process of the synthetic scenes, we generate $70$ cameras per object, render RGBD images to obtain initial point clouds, and reconstruct using the official 3D Gaussian Splatting implementation~\cite{kerbl20233d}.

\subsection{Evaluation Instances}

\begin{figure*}
    \centering
    \includegraphics[width=1\textwidth]{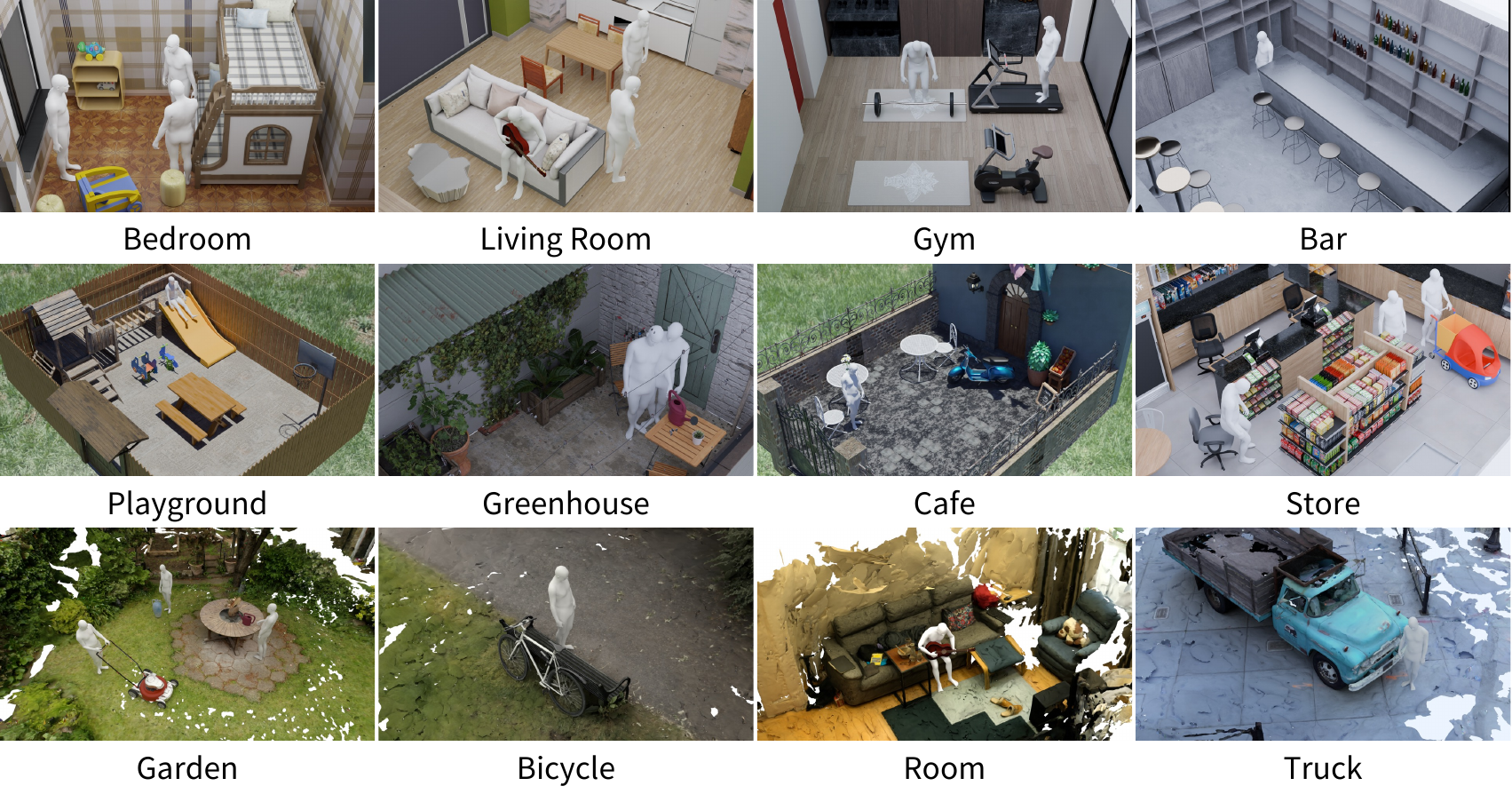}
    \caption{\textbf{Visualization of our \dataset dataset.}}
    \label{fig:dataset-visualization}
\end{figure*}

As shown in \cref{tab:dataset-statistics} and \cref{fig:dataset-visualization}, our \dataset dataset contains $22$ evaluation instances including $13$ static interaction instances and $9$ dynamic object interaction instances. Each interaction instance comprises a text prompt and initial state. The initial state typically features a standing pose with nearby objects. We adjust the standing pose for specific instances to ease the interaction video generation.

\section{Additional Experiment Details and Results}\label{sec:exp}

In this section, we provide additional; introduction to the experimental settings (\cref{sec:exp-setting}), implementation (\cref{sec:implementation,sec:camera-view-selection,sec:loss}), human studies (\cref{sec:human-study-detail}), and baseline comparisons (\cref{sec:comparison-detail}). We explain how our method is capable of synthesizing HSIs of varying lengths and present additional qualitative results of diverse long-term human-scene interactions (\cref{sec:long-term}). We show qualitative examples of our ablation study (\cref{sec:ablation-qualitative}). We also demonstrate our approach's flexibility by integrating various state-of-the-art video generation models beyond the model used in our main experiments (\cref{sec:other-model}).

\subsection{Experimental Settings}\label{sec:exp-setting}

For the static scenarios, we evaluate on $11$ static instances (excluding \textit{Bicycle} and \textit{Truck} scenes). We evaluate each scene with $5$ different seeds (as we consider a generative setting), yielding $55$ generated motion sequences for evaluation. For the dynamic scenarios, we evaluate on all $9$ dynamic object interaction instances. Similarly, we evaluate each scene with $5$ different seeds and this leads to $45$ motion sequences.

\subsection{Implementation Details}\label{sec:implementation}

We employ KLING image-to-video model v1.0~\cite{kling} for HSI video generation and SAM 2.0~\cite{ravi2024sam} for segmentation. For efficiency, we downsample the generated 153-frame videos to 51 frames for all experiments. For initialization, we employ a default standing pose with objects positioned near the human. We make adjustments for the initial pose and camera view for some instances. For comparative evaluation, we utilize the official implementations and pre-trained models of the baseline methods. Since all baseline approaches require spatial conditions for navigation, we provide them with the human/object trajectories generated by our method. We also supply the initialization configuration to these baselines, as they either require or can accommodate initial-state inputs.

The Adam optimizer~\cite{diederik2015adam} is utilized across all optimization stages: camera pose estimation, 4D HSI optimization, color net fine-tuning, and refinement. For camera pose estimation, we optimize $30$ iterations per frame with a learning rate of $0.001$. In 4D HSI optimization, we perform $300$ iterations per frame. During the initial $30$ iterations, the optimization of human poses is limited to root translation and global orientation. We use a learning rate of $0.01$, and the loss weights are set as $\lambda=0.1$, $\lambda_{\text{center}}=0.001$, and $\lambda_{\text{depth}}=0.001$. The color net is fine-tuned simultaneously every $5$ step during 4D HSI optimization with a learning rate of $0.00001$. For refinement, we optimize $1000$ iterations for the entire sequence with a learning rate of $0.05$ and physics loss weight $\lambda_{\text{physics}}=0.001$.

\begin{table*}[ht!]
    \centering
    \small
    \setlength{\tabcolsep}{3pt}
    \begin{tabular}{lll|ll}
        \toprule
        Scene Name & Scene Type & Source & Text Prompt & Objects \\
        \midrule
        \textit{Bedroom} & Indoor & TRUMANS~\cite{jiang2024scaling} & \textit{The person is sitting on the bed.} & Static  \\
        & & & \textit{The person is sitting on the windowsill.} & Static  \\
        & & & \textit{The person is leaning on the ladder.} & Static  \\
        \midrule
        \textit{Living Room} & Indoor & TRUMANS~\cite{jiang2024scaling} & \textit{The person is sitting on the table.} & Static  \\
        & & & \textit{The person is sitting on the sofa back.} & Static  \\
        & & & \textit{The person is playing guitar while sitting on the sofa.} & \textit{Guitar}  \\
        \midrule
        \textit{Gym} & Indoor & Asset Libraries & \textit{The person is running on the treadmill.} & Static  \\
        & & & \textit{The person is lifting weights.} & \textit{Barbell}  \\
        \midrule
        \textit{Bar} & Indoor & Asset Libraries & \textit{The person is leaning on the bar.} & Static  \\
        \midrule
        \textit{Playground} & Outdoor & Asset Libraries & \textit{The person is sliding down the slide.} & Static  \\
        \midrule
        \textit{Greenhouse} & Indoor & Asset Libraries & \textit{The person is sitting on the chair.} & Static  \\
        & & & \textit{The person is watering flowers with a watering can.} & \textit{Watering Can}  \\
        \midrule
        \textit{Cafe} & Outdoor & Asset Libraries & \textit{The person is sitting on the chair.} & Static  \\
        \midrule
        \textit{Store} & Indoor & Asset Libraries & \textit{The person is picking out snacks on the shelf.} & Static  \\
        & & & \textit{The person is sliding while sitting on the chair.} & \textit{Office Chair}  \\
        & & & \textit{The person is pushing shopping cart.} & \textit{Shopping Cart} \\
        \midrule
        \textit{Garden} & Outdoor & Mip-NeRF 360~\cite{barron2022mip} & \textit{The person is watering flowers with a watering can.} & \textit{Watering Can}  \\
        & & & \textit{The person is lifting a vase.} & \textit{Vase}  \\
        & & & \textit{The person is operating a lawn mower.} & \textit{Mower}  \\
        \midrule
        \textit{Bicycle} & Outdoor & Mip-NeRF 360~\cite{barron2022mip} & \textit{The person is sitting on the bench.} & Static  \\
        \midrule
        \textit{Room} & Indoor & Mip-NeRF 360~\cite{barron2022mip} & \textit{The person is playing guitar while sitting on the sofa.} & \textit{Guitar} \\
        \midrule
        \textit{Truck} & Outdoor & Tanks\&Temples~\cite{knapitsch2017tanks} & \textit{The person is cleaning the car.} & Static  \\
        \bottomrule
    \end{tabular}
    \caption{\textbf{Statistics of our \dataset dataset.}}
    \label{tab:dataset-statistics}
\end{table*}

\pagebreak
\clearpage
\clearpage

\subsection{Details on Initial Camera View Selection}\label{sec:camera-view-selection}

 Based on the assumption that camera movement remains minimal over brief durations (approximately 5 seconds), we address occlusion challenges by selecting an initial camera view $\mathbf{T}_0$ that maximizes visibility of both human subjects and dynamic objects in the first frame. We select camera from a predefined camera array distributed across concentric spherical surfaces with varying radii, all oriented toward the same sphere center. 

The sphere center's height is set as $\mathbf{c}^\text{Z} = h_\text{pelvis} + 0.1$ in a z-up coordinate system. For static scenes, its horizontal position equals the pelvis projection $\mathbf{p}_\text{pelvis}^\text{XoY}$.  For dynamic scenes, we identify the human's orientation direction $\mathbf{d}$ and create a ray $l$ from $\mathbf{p}_\text{pelvis}^\text{XoY}$ along $\mathbf{d}$. We then project dynamic object particles onto $l$ and set $\mathbf{c}^\text{XoY}$ as the midpoint between $\mathbf{p}_\text{pelvis}^\text{XoY}$ and the furthest projection.

For camera array generation, we place cameras on spherical surfaces at different radii, using only the hemisphere facing the human but avoiding direct frontal views. All cameras are oriented toward the sphere center. This setup balances complete human capture while reducing depth ambiguity in the generated video.

For each candidate camera, we render two depth maps: $D$ (entire scene) and $D_\mathcal{H}$ (human only). Non-occluded areas are defined as $\Omega=\big\{(u,v)\big|\vert D(u,v)- D_\mathcal{H}(u,v)\vert<\epsilon\big\}$. We project the first 22 SMPL-X joints~\cite{pavlakos2019expressive} onto the image plane and count visible joints as $N_\text{visible}$. The camera with maximum $N_\text{visible}$ is selected. With equal visibility, we choose the camera closest to the equatorial plane for better accuracy for height reconstruction.

\subsection{Details on Physics Loss}\label{sec:loss}

We leverage a hand-object contact loss to encourage hand-object contact when they are in close proximity after the optimization process described in \cref{sec:hsi-optimization}. We first define the contact set $\mathcal{C}_t$ for each frame as:
\begin{equation}
    \mathcal{C}_t=\Big\{(\mathbf{v}_i, \boldsymbol{\mu}_p)\Big|\big\Vert\mathbf{v}_i-\boldsymbol{\mu}_p\big\Vert_2<\epsilon,\mathbf{v}_i\in\mathbf{H}_t,\boldsymbol{\mu}_p\in\boldsymbol{\mu}_{\mathcal{O}}^t\Big\},
\end{equation}

where $\mathbf{H}_t$ represents SMPL-X~\cite{pavlakos2019expressive} hand vertices and $\boldsymbol{\mu}_{\mathcal{O}}^t$ denotes Gaussian object particle positions in frame $t$. The frame-wise hand-object contact loss is then defined as:
\begin{equation}
    \mathcal{L}_{\text{contact}}^t = \frac{1}{\#\mathcal{C}_t}\min_{\mathbf{d}}\Big\{\mathbf{d}=\big\Vert\mathbf{v}_i-\boldsymbol{\mu}_p\big\Vert_2\Big|(\mathbf{v}_i, \boldsymbol{\mu}_p)\in\mathcal{C}_t\Big\},
\end{equation}
which we only apply when $\#\mathcal{C}_t>0$.

Additionally, since VPoser~\cite{pavlakos2019expressive} as a human pose prior does not inherently ensure sequence smoothness after refinement, we apply an additional smoothness loss between adjacent frames:
\begin{equation}
\mathcal{L}_{\text{smooth}}=\frac{1}{T-1}\sum_{t=0}^{T-1}\big\Vert\mathbf{\Theta}_t-\mathbf{\Theta}_{t+1}\big\Vert_2.
\end{equation}

The overall physics loss is defined as:
\begin{equation}
\mathcal{L}_{\text{physics}}=\lambda_{\text{contact}}\frac{1}{T}\sum_{t=0}^T\mathcal{L}_{\text{contact}}^t+\lambda_{\text{smooth}}\mathcal{L}_{\text{smooth}},
\end{equation}
where we set $\lambda_{\text{contact}}=1$ and use $\lambda_{\text{smooth}}=0.3$ for static scenarios and $\lambda_{\text{smooth}}=0.1$ for dynamic scenarios.

\begin{figure}[t!]
    \centering
    \includegraphics[width=1\linewidth]{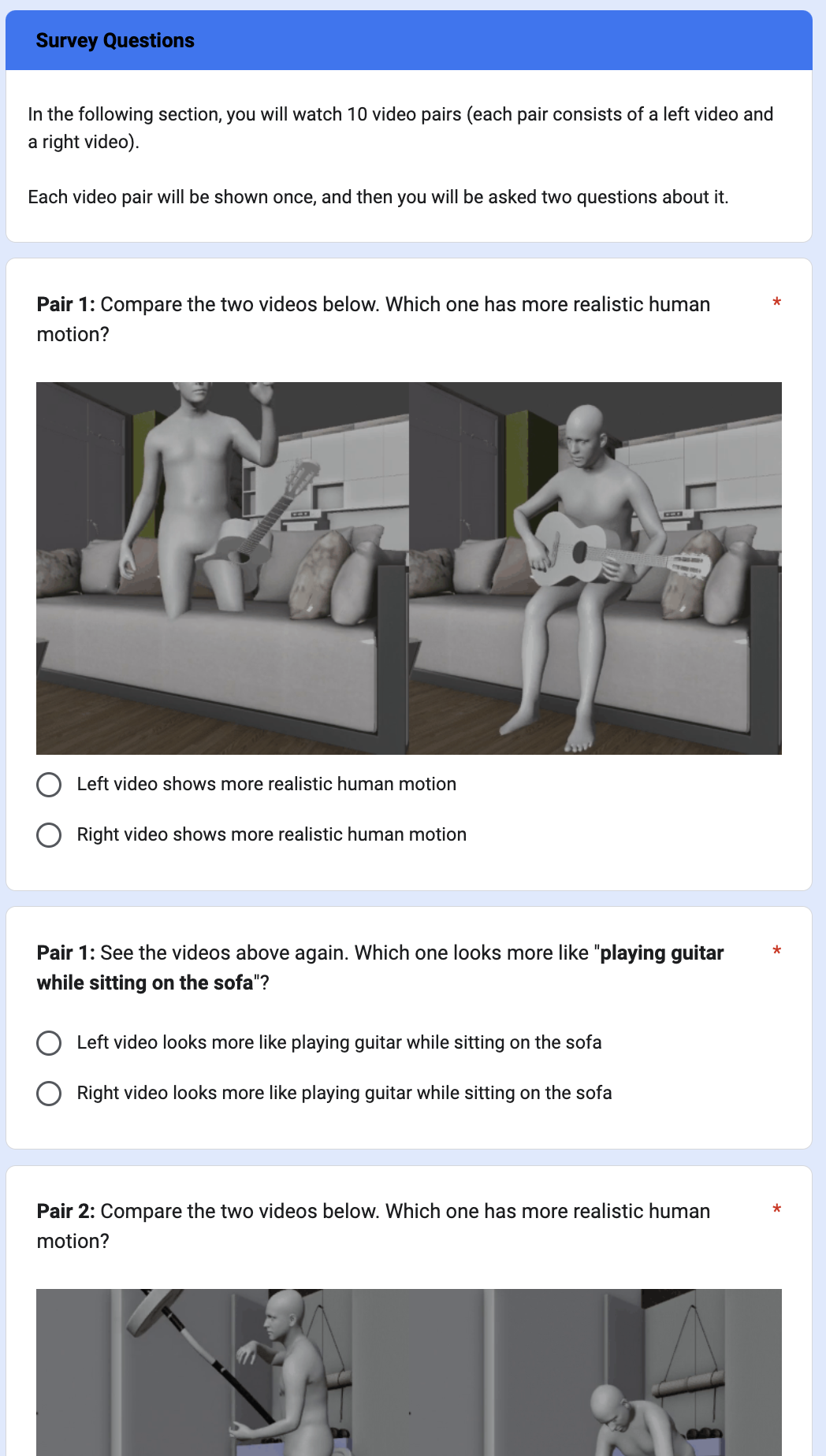}
    \caption{Screenshot of the human study interface.}
    \label{fig:human-study}
\end{figure}

\begin{figure*}[t!]
    \centering
    \begin{subfigure}{\linewidth}
        \centering
        \includegraphics[width=\linewidth]{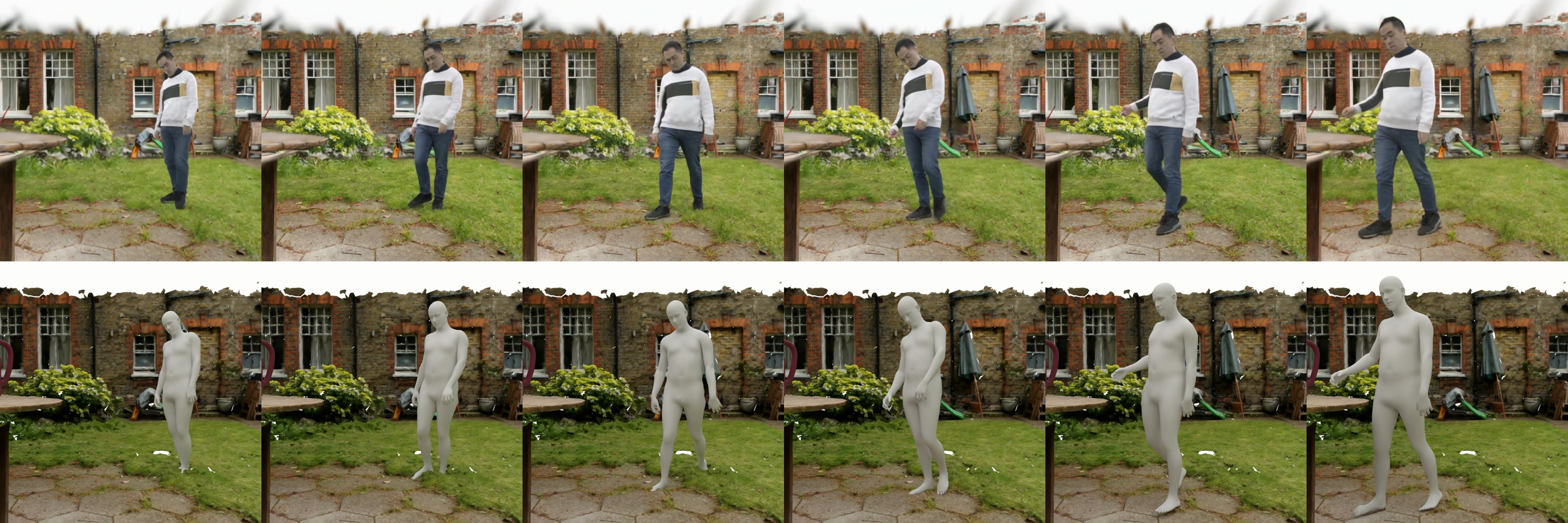}
        \caption{``Walking forward to the table.''}
    \end{subfigure}
    \\
    \begin{subfigure}{\linewidth}
        \centering
        \includegraphics[width=\linewidth]{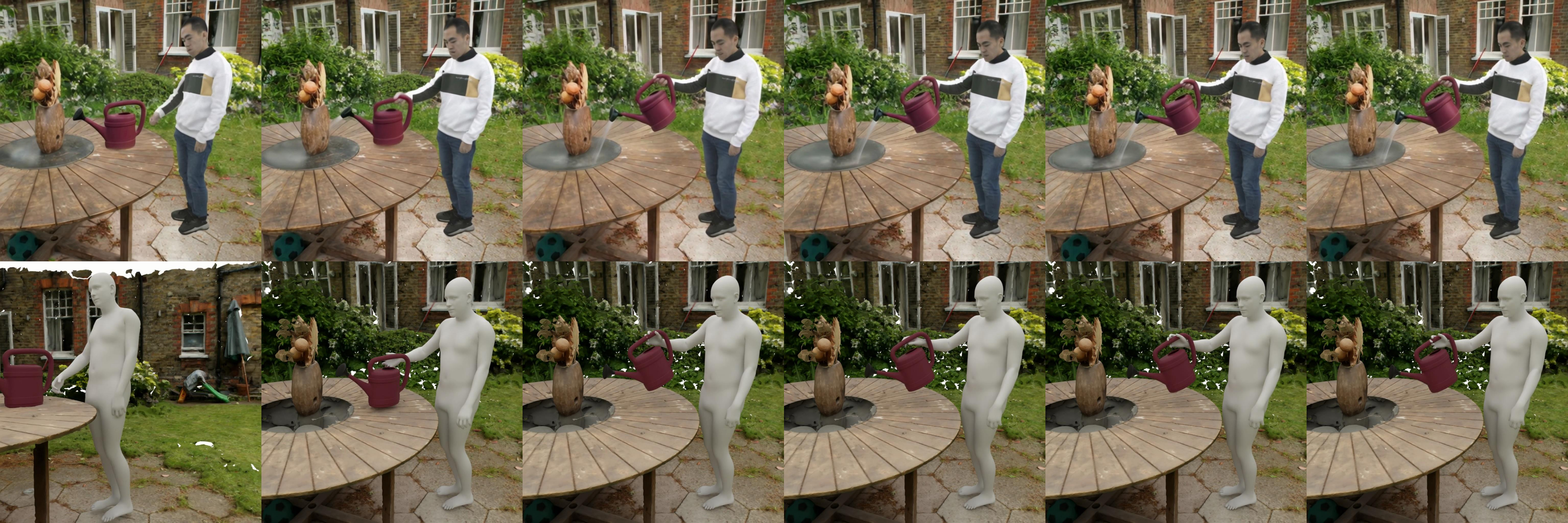}
        \caption{``Watering flowers with a watering can.''}
    \end{subfigure}
    \\
    \begin{subfigure}{\linewidth}
        \centering
        \includegraphics[width=\linewidth]{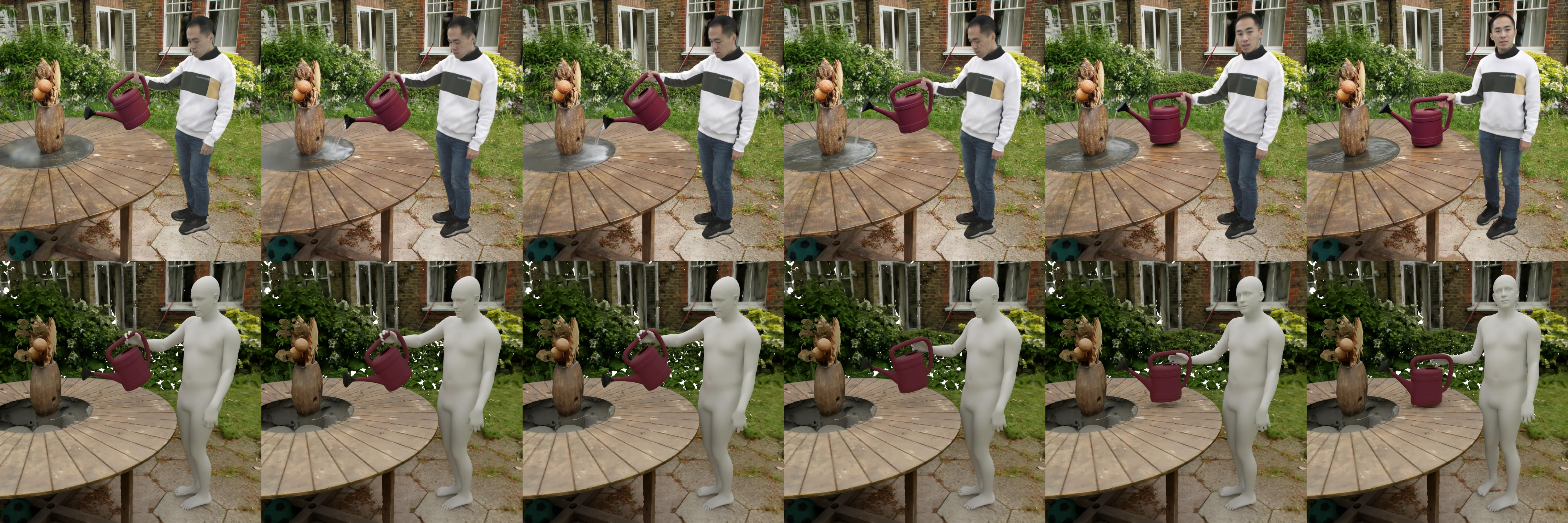}
        \caption{``Putting down the watering can.''}
    \end{subfigure}
    \caption{\textbf{Example of long-term human-scene interaction sequence.} \model generates this long interaction sequence from three 5-second HSI videos.}
    \label{fig:long-term-1}
    \vspace{-12pt}
\end{figure*}

\begin{figure*}[t!]
    \centering
    \begin{subfigure}{\linewidth}
        \centering
        \includegraphics[width=\linewidth]{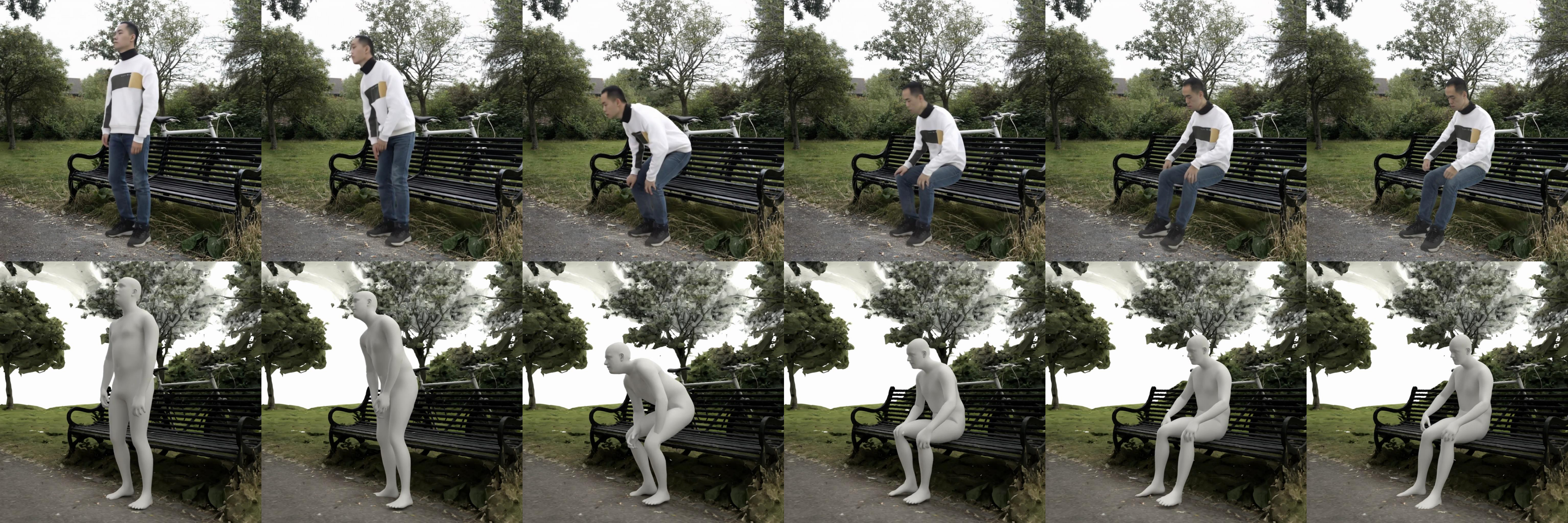}
        \caption{``Sitting on the bench.''}
    \end{subfigure}
    \\
    \begin{subfigure}{\linewidth}
        \centering
        \includegraphics[width=\linewidth]{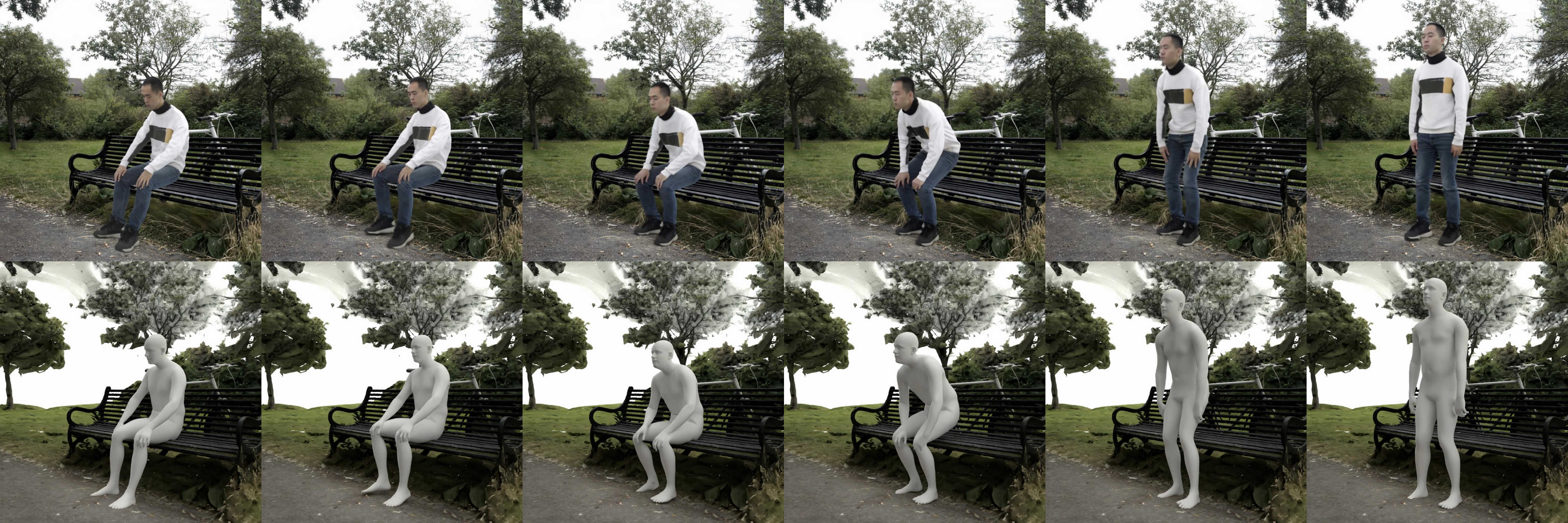}
        \caption{``Standing up.''}
    \end{subfigure}
    \caption{\textbf{Example of long-term human-scene interaction sequence.} \model generates this long interaction sequence from two 5-second HSI videos (sitting down, standing up) and one 10-second HSI video (practicing Tai Chi).}
    \label{fig:long-term-2-1}
    \vspace{-12pt}
\end{figure*}

\begin{figure*}[t!]
    \ContinuedFloat
    \renewcommand{\thefigure}{S\arabic{figure}}
    \setcounter{figure}{5}
    \centering
    \begin{subfigure}{\linewidth}
        \centering
        \includegraphics[width=\linewidth]{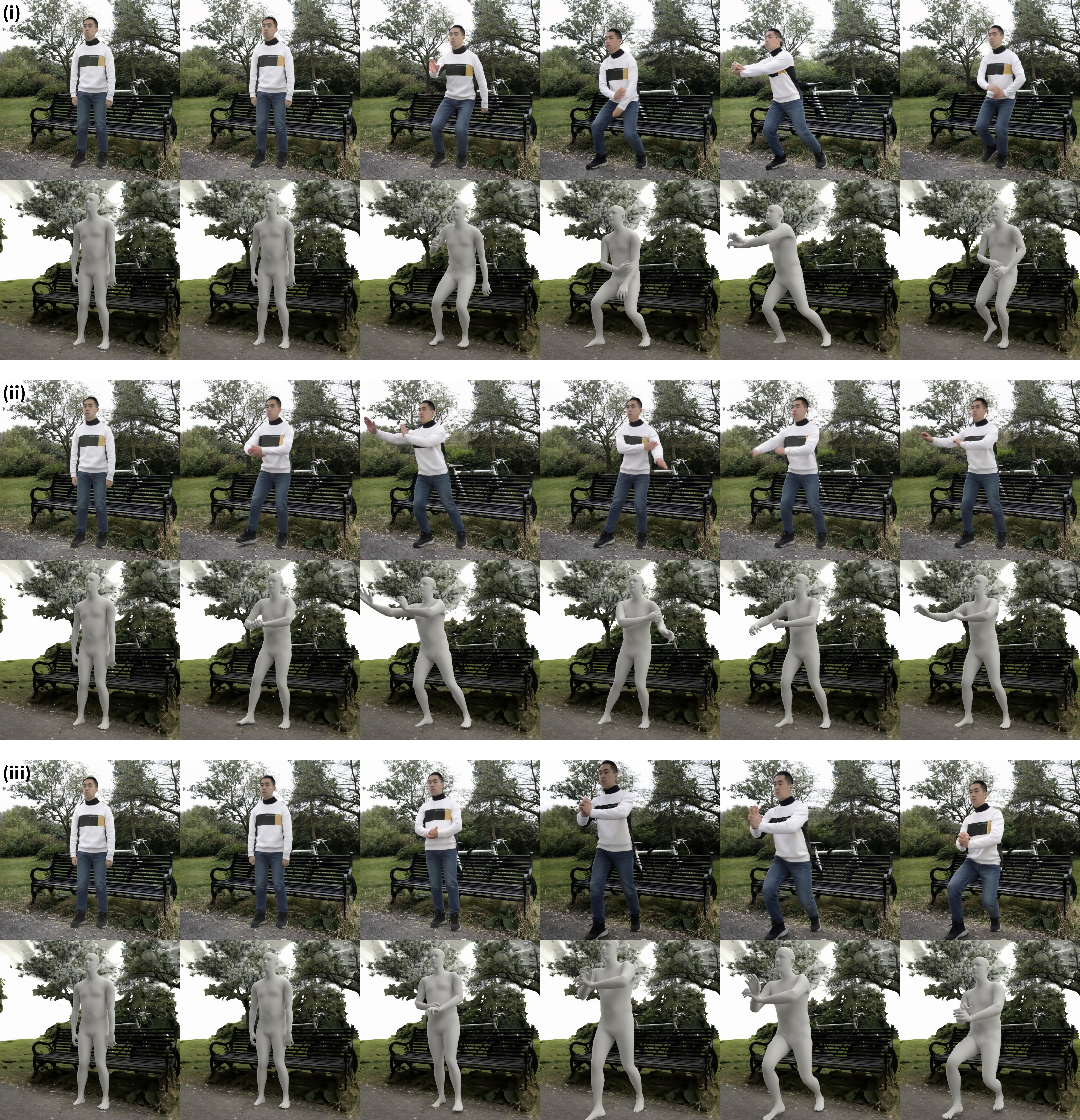}
        \caption{``Practicing Tai Chi.''}
    \end{subfigure}
    \caption{\textbf{Example of long-term human-scene interaction sequence.} \model generates this long interaction sequence from two 5-second HSI videos (sitting down, standing up) and one 10-second HSI video (practicing Tai Chi). We show multiple variations of practicing Tai Chi generated by \model, demonstrating its diverse generation capability.}
    \label{fig:long-term-2-2}
\end{figure*}

\subsection{Human Study Details}\label{sec:human-study-detail}

We recruit $400$ participants via the Prolific platform. The participants are divided into $4$ groups, each of which includes $100$ participants. Each participant is shown two side-by-side videos and forced to choose one from them. One of the videos is ours, and the other is generated by a baseline method. The left-right order is randomized. Each participant is shown $10$ pairs of videos. For each pair of videos, the participant is asked two questions: (1) to choose the video that is higher quality, and (2) to choose the video that is better aligned with the text prompt. We show a screenshot of the human study interface in \cref{fig:human-study}.

\subsection{Comparison Details}\label{sec:comparison-detail}
\vspace{-4pt}

In each generation process, our \model outputs a 5-second HSI sequence at 10 fps. All baseline methods output sequences at 30 fps, which we downsample to 10 fps for comparison. For a fair comparison, we do not apply refinement to our outputs or baseline results. Since TRUMANS~\cite{jiang2024scaling} and LINGO~\cite{jiang2024autonomous} require occupancy grid inputs, we convert meshes from both synthetic scenes and real scenes (via SuGaR~\cite{guedon2024sugar}) into occupancy grids, yielding water-tight meshes as a byproduct. We use these water-tight meshes for penetration metric calculations, as they enable proper inside-outside definition necessary for Signed Distance Field (SDF) computation.

\subsection{Long-Term Interaction Synthesis}\label{sec:long-term}

As shown in \cref{fig:teaser} (first walking forward, then watering flowers), our method inherently supports the generation of long-term HSI sequences. While KLING image-to-video mode is limited to generating videos under $10$ seconds and only allows $5$-second extensions per generation, we overcome this limitation by using the last frame of each generated HSI sequence as the initial state for the subsequent generation process. By repeating the entire generation process described in~\cref{sec:method}, we can synthesize longer sequences. This approach ensures consistent quality across each generated sequence while enabling flexible control over individual video clips. For dynamic object interactions, the depth regularization term varies between different 5-second video clips, making the constant depth assumption within each sequence reasonable.

\cref{fig:long-term-1} presents a 15-second interaction sequence where the character walks forward, waters flowers, and puts down the watering can. We achieve this extended interaction by generating and reconstructing three HSI videos into a cohesive 4D sequence.

\cref{fig:long-term-2-1} and \cref{fig:long-term-2-2} showcase another 20-second sequence where the character sits on a bench, stands up, and performs Tai Chi. Leveraging diverse video generation capabilities, \model can create multiple variations of this interaction, as demonstrated by the alternative Tai Chi sequences (ii) and (iii) in \cref{fig:long-term-2-2}.

\begin{figure*}[t!]
    \centering
    \includegraphics[width=1\textwidth]{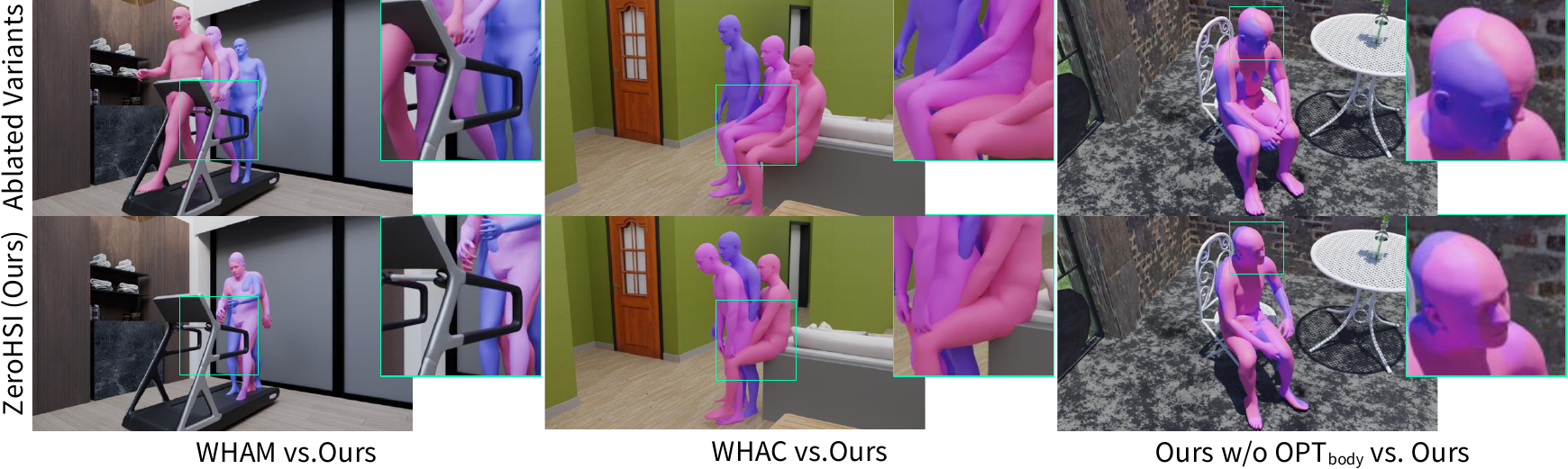}
    \caption{\textbf{Qualitative results of ablation study on our optimization-based HSI motion reconstruction.} Our full method reconstructs root translation more accurately than WHAM~\cite{shin2024wham} and WHAC~\cite{yin2024whac} while achieving smoother results than \model w/o OPT$_\text{body}$.}
    \label{fig:results-ablation}
    \vspace{-4pt}
\end{figure*}

\begin{figure*}[t!]
    \centering
    \includegraphics[width=1\textwidth]{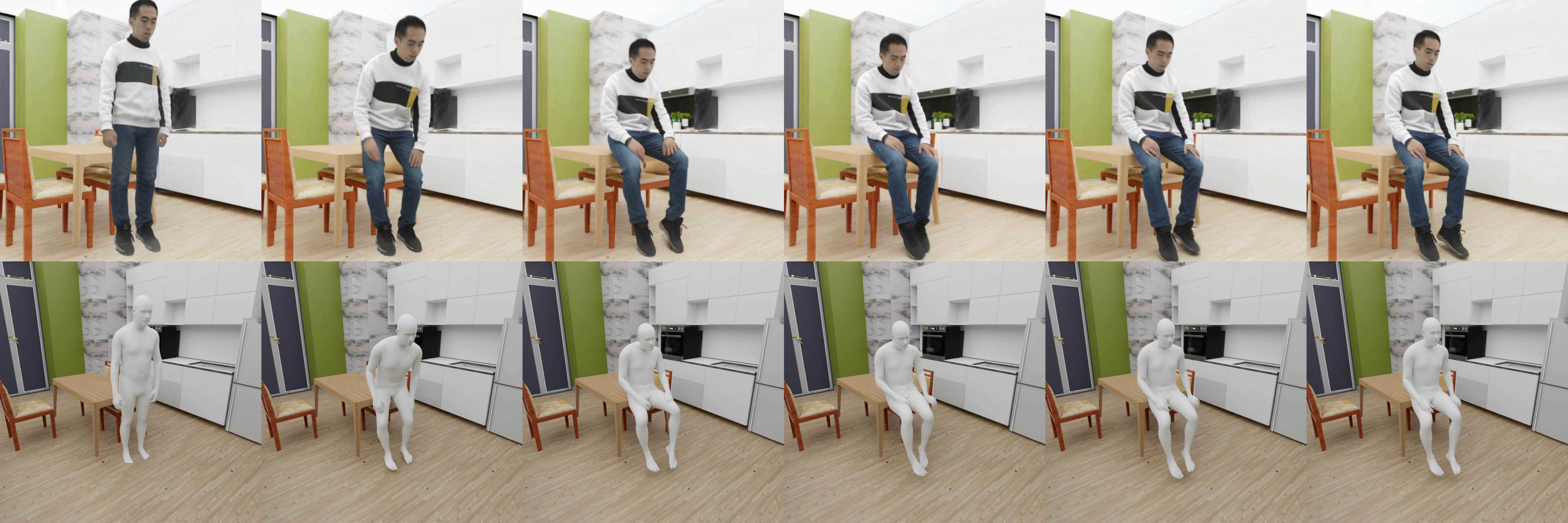}
    \caption{\textbf{HSI generated with KLING 1.5~\cite{kling}.} Text prompt: ``Sitting on the table.''}
    \label{fig:kling-1-5}
\end{figure*}

\begin{figure*}[t!]
    \centering
    \includegraphics[width=1\textwidth]{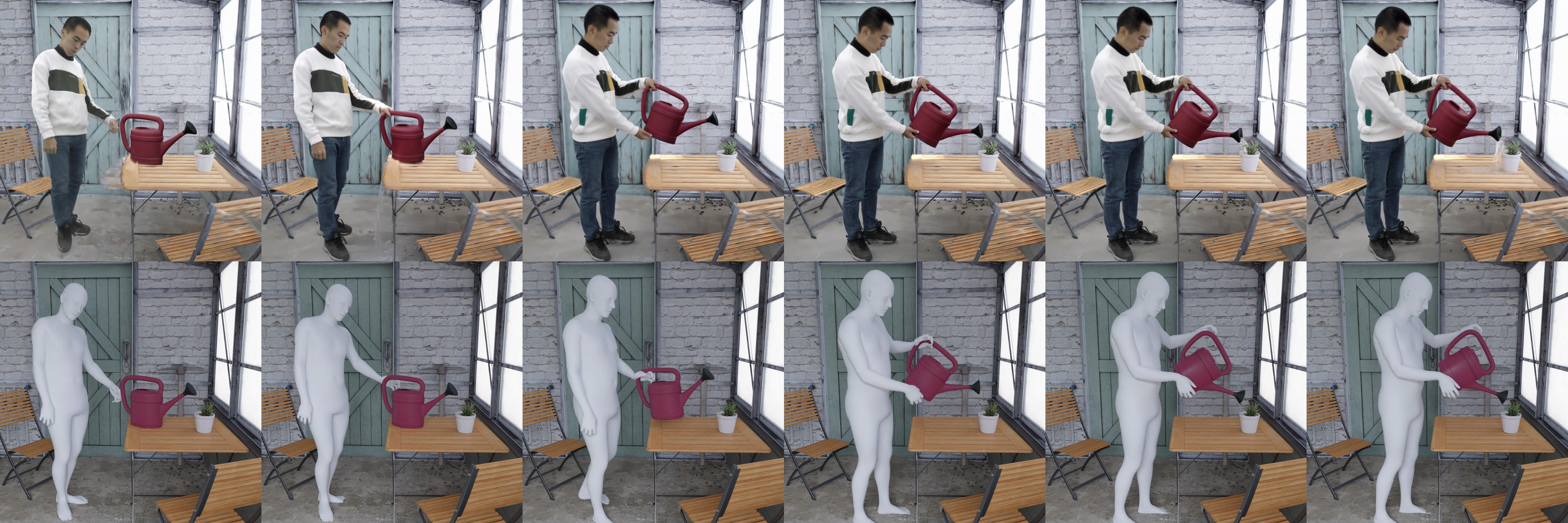}
    \caption{\textbf{HSI generated with KLING 1.6~\cite{kling}.} Text prompt: ``Watering flowers with a watering can.''}
    \label{fig:kling-1-6}
\end{figure*}

\begin{figure*}[t!]
    \centering
    \includegraphics[width=1\textwidth]{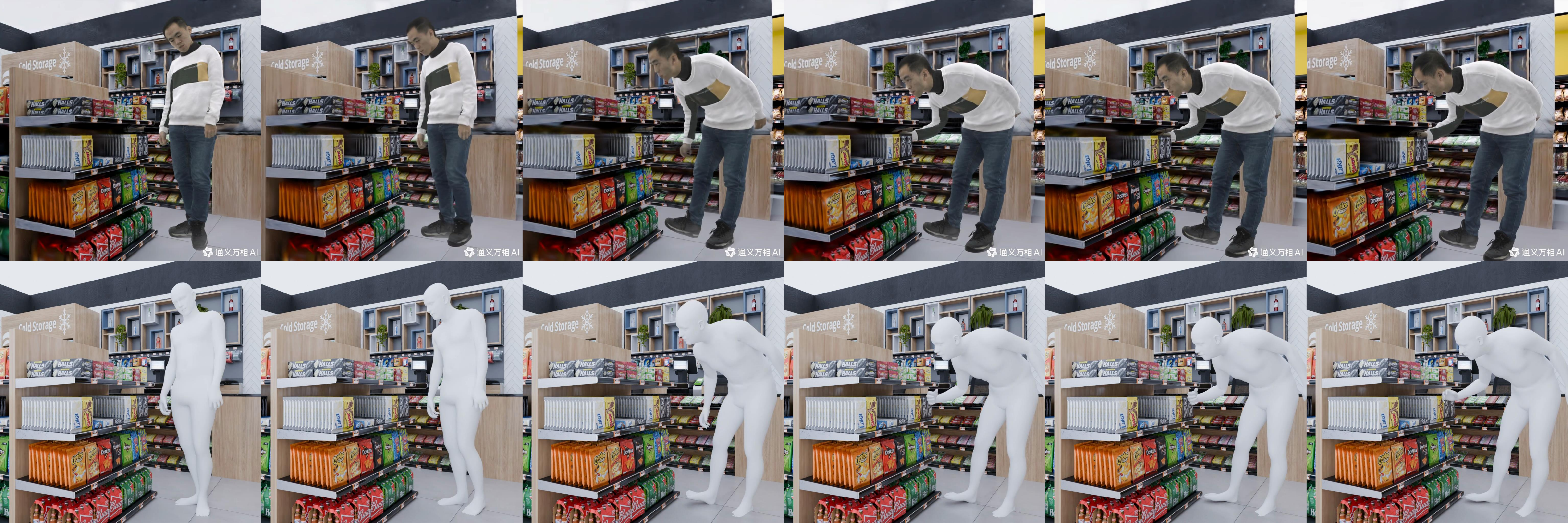}
    \caption{\textbf{HSI generated with Wan 2.1~\cite{wan2.1}.} Text prompt: ``Picking out snacks on the shelf.''}
    \label{fig:wan-2-1}
\end{figure*}

\begin{figure*}[t!]
    \centering
    \includegraphics[width=1\textwidth]{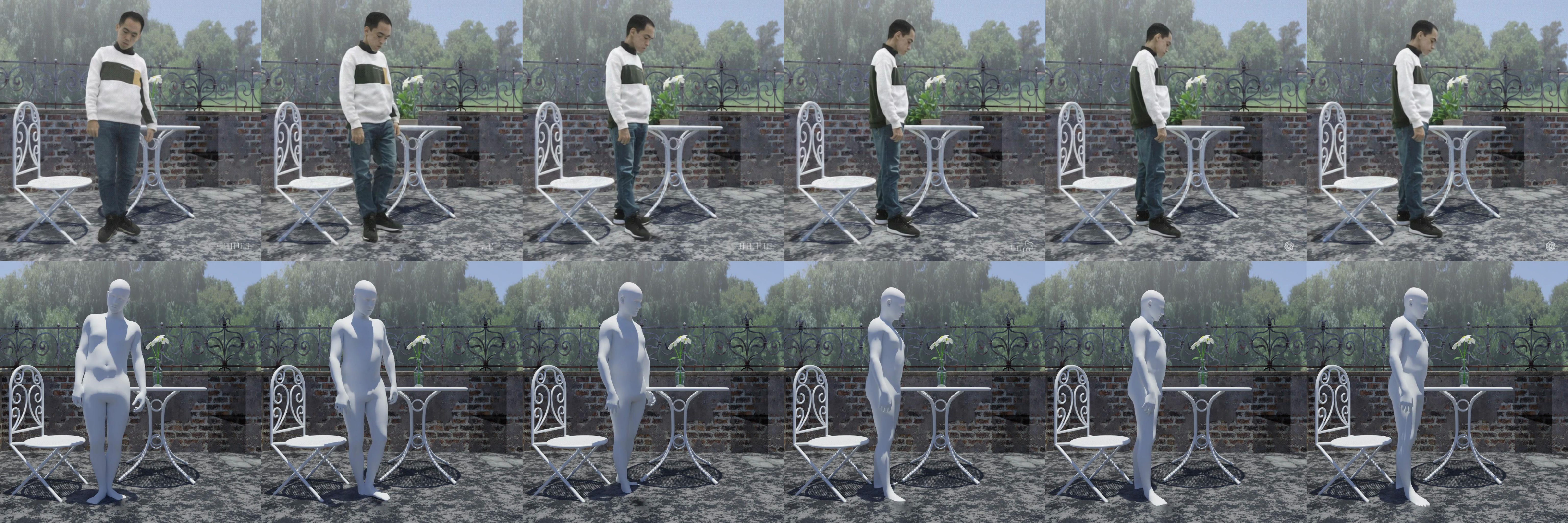}
    \caption{\textbf{HSI generated with Sora~\cite{brooks2024video}.}}
    \label{fig:sora}
\end{figure*}

\begin{figure*}[t!]
    \centering
    \begin{subfigure}{0.49\linewidth}
        \centering
        \includegraphics[width=\linewidth]{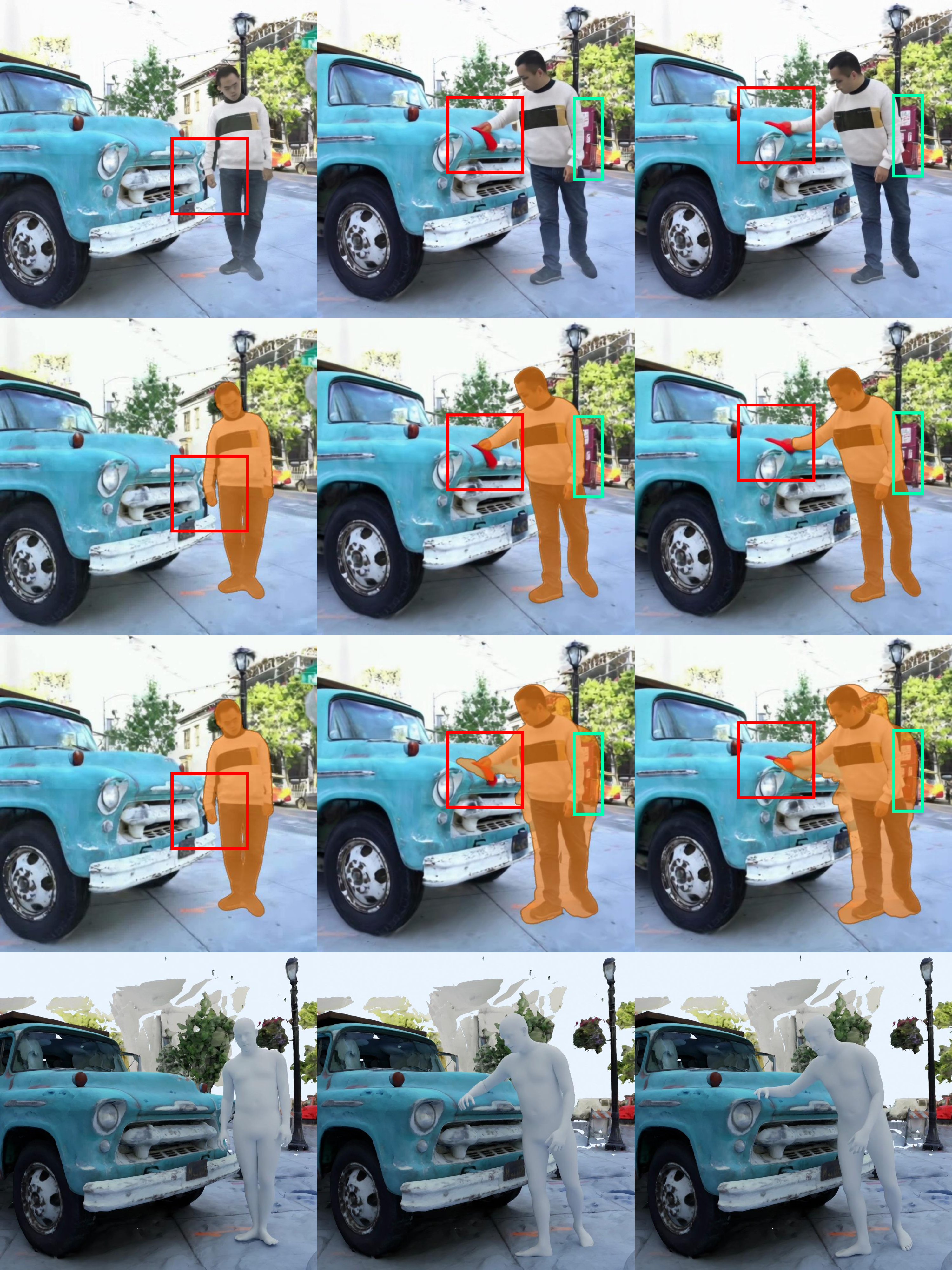}
        \caption{A \textcolor{red}{cloth} appears in hand during car wiping; a \textcolor{MyGreen}{mailbox} appears behind as the character leans forward.}
    \end{subfigure}
    \begin{subfigure}{0.49\linewidth}
        \centering
        \includegraphics[width=\linewidth]{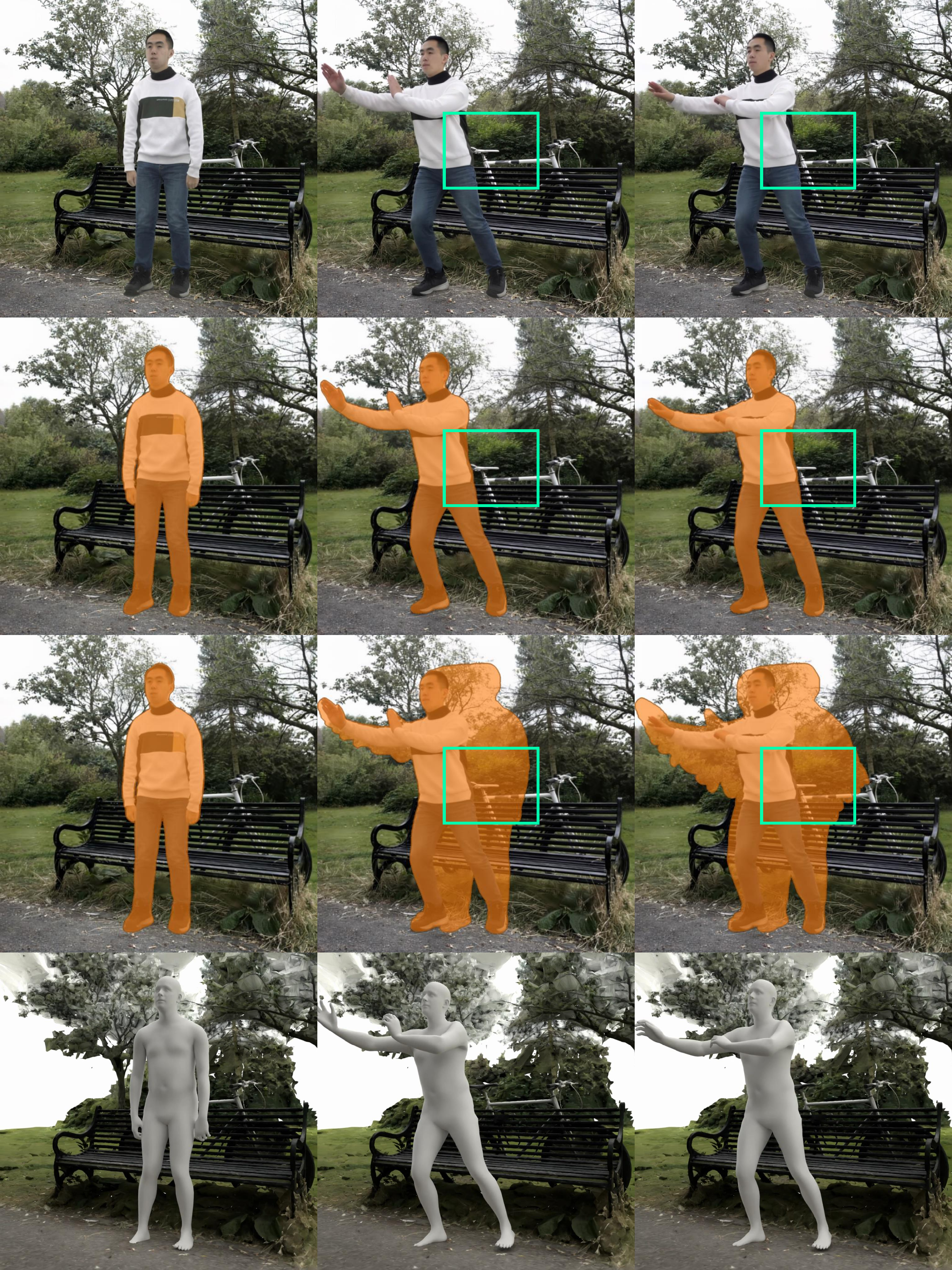}
        \caption{The \textcolor{MyGreen}{bicycle seat} appears white in the generated video while it is black in the real scene.}
    \end{subfigure}
    \vspace{-4pt}
    \caption{\textbf{Example of incorrect contents.} The first row shows frames containing unwanted contents. The second row shows results using only static background masks, while the third row shows our aggregated dynamic foreground masking approach. The fourth row presents our final reconstruction results.}
    \label{fig:incorrect-contents}
    \vspace{-12pt}
\end{figure*}

\begin{figure*}[t!]
    \centering
    \begin{subfigure}{0.49\linewidth}
        \centering
        \includegraphics[width=\linewidth]{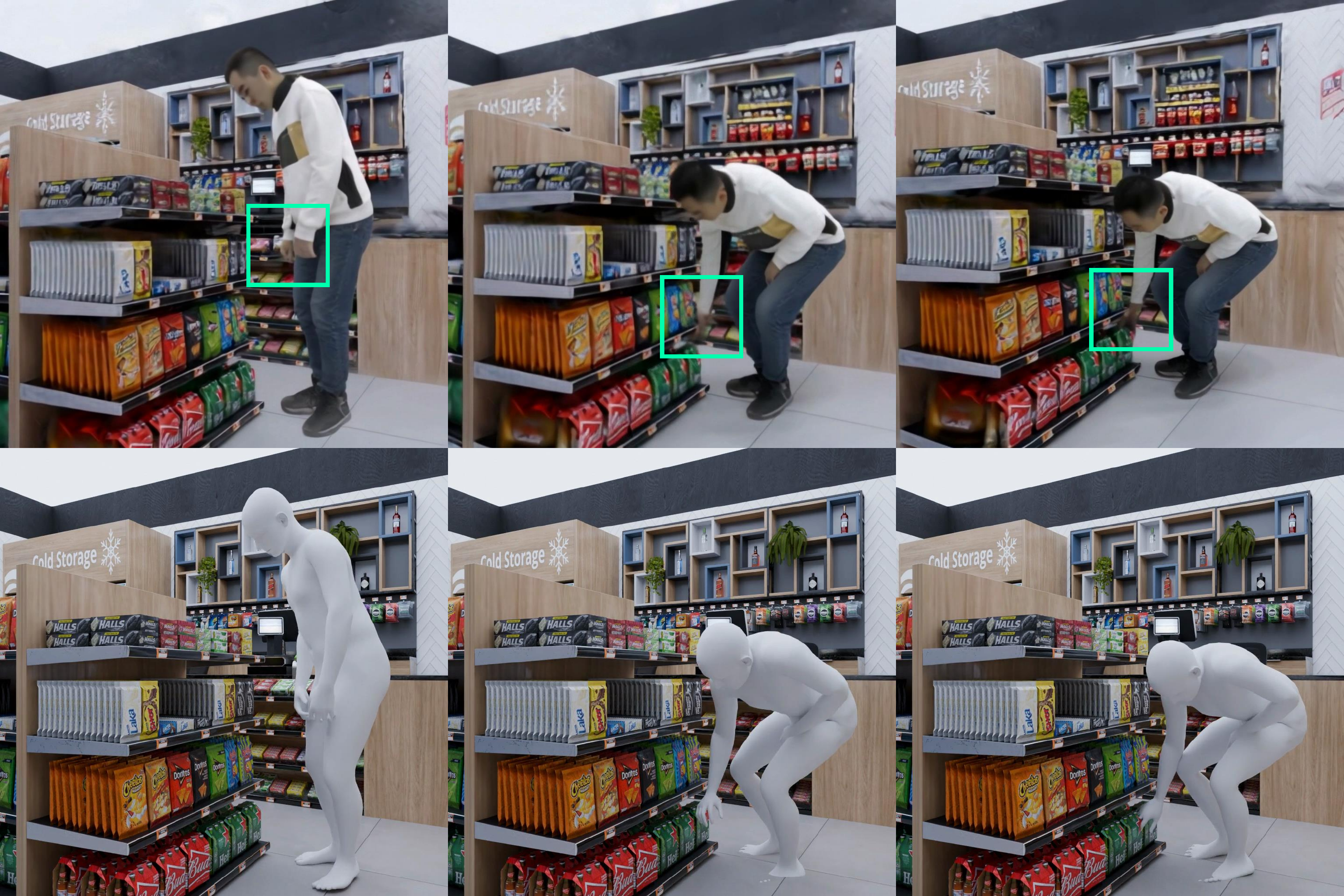}
        \caption{As the character squats down, their \textcolor{MyGreen}{right hand} fades and blends into the background shelf.}
    \end{subfigure}
    \begin{subfigure}{0.49\linewidth}
        \centering
        \includegraphics[width=\linewidth]{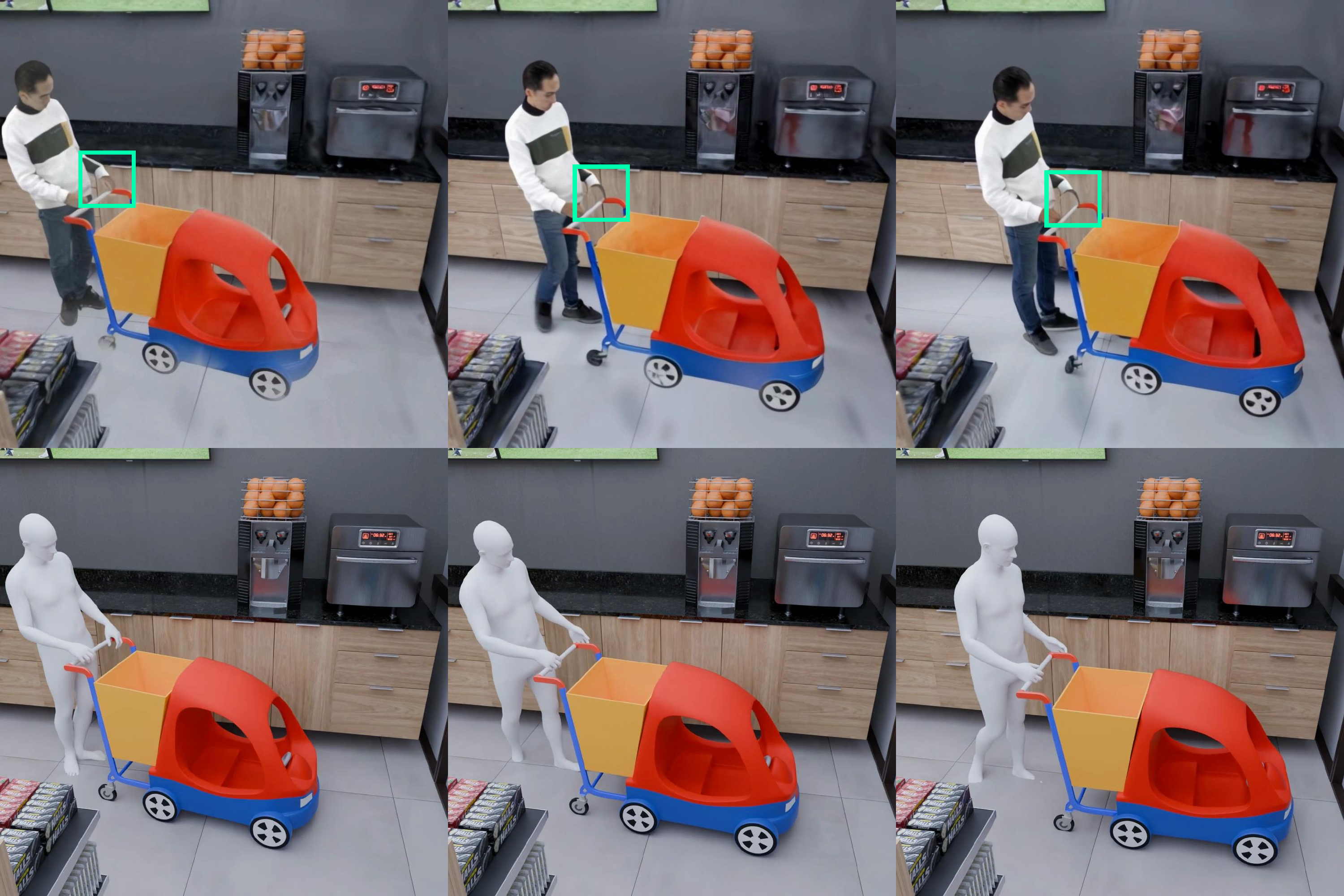}
        \caption{As the character walks forward, their \textcolor{MyGreen}{left hand} fades and blends into the background cabinet.}
    \end{subfigure}
    \caption{\textbf{Example of body part disappearance.} The first row shows frames of  body part disappearance. The second row presents our final reconstruction results.}
    \label{fig:body-part-disappear}
    \vspace{-12pt}
\end{figure*}

\begin{figure*}
    \centering
    \includegraphics[width=1\textwidth]{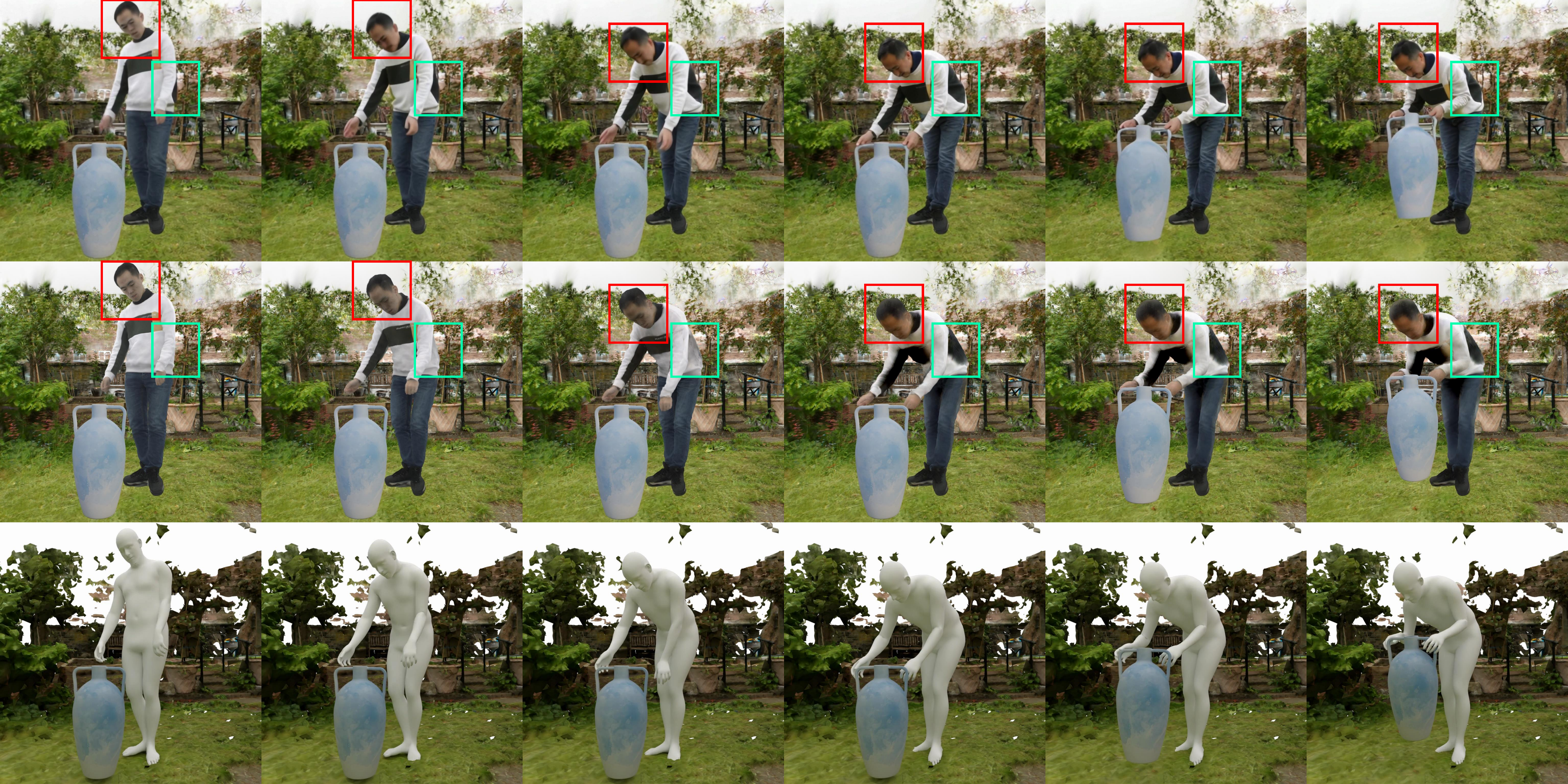}
    \caption{\textbf{Example of human appearance change.} The first row shows frames of human appearance change. The character's \textcolor{red}{skin tone and hairstyle} change while leaning forward; \textcolor{MyGreen}{black patterns} incorrectly shift from the sleeves onto the torso. The second row shows gaussian rendering results after optimization. The third row presents our final reconstruction results.}
    \label{fig:appearance-change}
    \vspace{-12pt}
\end{figure*}

\vspace{-2pt}
\subsection{Qualitative Results of Ablations}\label{sec:ablation-qualitative}
\vspace{-2pt}

We show qualitative results of our ablation study on our per-frame optimization approach. The examples in \cref{fig:results-ablation} reveal that both WHAM~\cite{shin2024wham} and WHAC~\cite{yin2024whac} fail to estimate correct global translation, resulting in penetrations between human and scene. Meanwhile, rendered results across three consecutive frames show our optimization-based method achieving smoother motions compared to ``\model w/o OPT$_\text{body}$''. These results further demonstrate the effectiveness and technical contribution of our per-frame optimization approach on reconstructing 4D HSIs from 2D videos. 

\subsection{Results with Other Video Generation Models}\label{sec:other-model}

We use off-the-shelf video generation model to ensure \model maintains compatibility with various existing techniques and remains adaptable to future advancements in video generation. To evaluate \model's adaptability across different models, we test it with other state-of-the-art image-to-video models: KLING 1.5~\cite{kling}, KLING 1.6~\cite{kling}, Wan 2.1~\cite{wan2.1}, and Sora~\cite{brooks2024video}.

Qualitative results in \cref{fig:kling-1-5,fig:kling-1-6,fig:wan-2-1,fig:sora} demonstrate \model's effectiveness across different open-source video generation models and proprietary models with external APIs. \model performs well with newer versions of KLING in both static (``sitting on table'') and dynamic (``watering flowers'') scenarios. With Wan 2.1, \model successfully generates plausible interaction motions that follow the provided text prompt. We observe that Sora struggle to generate videos aligning well with the given text. However, when we examine it without textual input, our \model still generates meaningful 4D HSIs of ``watching the vase on the table.'' 

\section{Example of Low-Quality Video}\label{sec:low-quality-video}

In this section, we demonstrate the robustness of our method by presenting challenging examples of low-quality video inputs. Specifically, we incorporate techniques for handling incorrect content in camera pose estimation, as well as dealing with body part disappearance and human appearance changes during HSI reconstruction. For each issue, we provide representative examples and show how our method successfully addresses these challenges through the reconstruction results.

\subsection{Incorrect Contents}\label{sec:incorrect-contents}

Video generation models produce incorrect contents in regions initially occluded by humans or objects that later become visible, and we use the aggregating dynamic foreground masks from frames 0 to $t$ to eliminate incorrect contents. \cref{fig:incorrect-contents} presents two examples of such incorrect generated content. In the first example, a cloth appears in the character's hand during car wiping, and a mailbox appears behind the character as they lean forward. In the second example, the color of the bicycle seat behind the human is generated incorrectly (black in the real scene but white in the generated video).

We compare two approaches: directly using the static background mask in the current frame (second row) versus using the aggregating dynamic foreground masks from frames 0 to $t$ (third row). Using only the static background mask in the current frame fails to mask out all unwanted contents, while using the complementary set of the aggregated dynamic foreground mask successfully eliminates all incorrect contents. The plausible reconstruction results in the last row demonstrate the effectiveness of our design.

\subsection{Body Part Disappearance}\label{sec:body-part-disappearance}

\cref{fig:body-part-disappear} demonstrates body part disappearance in both static and dynamic scenarios, where characters' hands gradually fade and blend into the background. Our per-frame optimization method produces natural and cohesive reconstructed HSIs (second row), demonstrating its robustness to such issues.

\subsection{Change of Human Appearance}\label{sec:appearance-change}

Video generation models often produce inconsistent human appearance. To address this, we fine-tune the color net defined in \cref{eq:color-net} during optimization using the photometric loss in each frame (\cref{eq:photometric-loss}). \cref{fig:appearance-change} shows an example where a character's skin tone, hairstyle, and clothing patterns change incorrectly. The second row displays gaussian rendering results after per-frame optimization, demonstrating how our approach successfully corrects appearance inconsistencies by gradually fine-tuning the color net. The final row shows plausible reconstruction results, validating our design's effectiveness.

\section{Algorithms}\label{sec:alg}

We summarize \model and show the overall algorithm in Alg.~\ref{alg:model}. 

\begin{algorithm*}[t!]
    \caption{\model: Zero-shot Human-Scene Interaction Generation}
    \label{alg:model}
    \begin{algorithmic}[1]
    \State\textbf{Input:}
        \State Scene Gaussians $\mathcal{G}_\mathcal{S}$, Object Gaussians $\mathcal{G}_\mathcal{O}$
        \State Initial human pose $\mathcal{M}_0=(\mathbf{r}_0,\boldsymbol{\phi}_0,\mathbf{\Theta}_0)$
        \State Initial object pose $\mathbf{P}_0$, Initial camera pose $\mathbf{T}_0$
        \State Text prompt $c$ describing the interaction
    
    \Function{GenerateHSIVideo}{}
        \State $\mathcal{G}_\mathcal{H}^0 \leftarrow \mathcal{A}(\mathbf{r}_0,\boldsymbol{\phi}_0,\mathbf{\Theta}_0;\mathbf{T}_0)$ \Comment{Initialize human Gaussians}
        \State $\mathcal{G}_\mathcal{O}^0 \leftarrow \mathcal{G}_\mathcal{O}(\mathbf{P}_0)$ \Comment{Transform object Gaussians}
        \State $\mathbf{I}_0 \leftarrow \mathcal{R}(\mathcal{G}_\mathcal{H}^0,\mathcal{G}_\mathcal{O}^0,\mathcal{G}_\mathcal{S};\mathbf{T}_0)$ \Comment{Render initial frame}
        \State $\{\mathbf{I}_t\}_{t=0}^T \leftarrow \text{VideoGen}(\mathbf{I}_0, c)$ \Comment{We use KLING}
        \State $\{\mathbf{M}_\mathcal{H}^t,\mathbf{M}_\mathcal{O}^t\}_{t=0}^T \leftarrow \text{SAM2}(\{\mathbf{I}_t\}_{t=0}^T)$ \Comment{Segment video}
    \State\textbf{return} $\{\mathbf{I}_t,\mathbf{M}_\mathcal{H}^t,\mathbf{M}_\mathcal{O}^t\}_{t=0}^T$
    \EndFunction
    
    \Function{Reconstruct4DHSI}{}
        \For{$t = 1$ to $T$}
            \State $\mathbf{T}_* \leftarrow \arg\min_{\mathbf{T}}\mathcal{L}_2(\mathcal{R}(\mathcal{G}_\mathcal{S}(\mathbf{T});\mathbf{T}_{t-1})\odot\mathbf{M}_t, \mathbf{I}_t\odot\mathbf{M}_t)$ \Comment{Estimate camera pose}
            \State $\mathbf{T}_t \leftarrow \mathbf{T}_*^{-1}\mathbf{T}_{t-1}$
            \State Initialize $\mathbf{r}_t,\boldsymbol{\phi}_t$ from previous frame \Comment{Initialize global transform}
            \State Initialize $\mathbf{\Theta}_t$ from pose estimation model
            \State $\mathcal{M}_t,\mathbf{P}_t \leftarrow \arg\min(\mathcal{L}_{\text{rgb}} + \lambda_{\text{center}}\mathcal{L}_{\text{center}} + \lambda_{\text{depth}}\mathcal{L}_{\text{depth}})$ \Comment{Optimize human and object poses}
        \EndFor
        \State\textbf{return} $\{(\mathcal{M}_t,\mathbf{P}_t)\}_{t=1}^T$
    \EndFunction

    \Function{Refinement}{}        
        \For{$t = 1$ to $T$}
            \State $\hat{J}_t \leftarrow \text{SMPL-X}(\mathcal{M}_t)$ \Comment{Get reference joints}
            \State $\mathcal{L}_{\text{fit}}^t \leftarrow \mathcal{L}_2\Big(\hat{J}_t, J_t\big(\mathbf{r}_t, \boldsymbol{\phi}_t, \mathcal{D}(\mathbf{z}_t)\big)\Big)$ \Comment{Compute fitting loss for frame $t$}
        \EndFor
        \State $\mathcal{L}_{\text{physics}} \leftarrow \text{CalculatePhysicsLoss}(\{\mathbf{r}_t, \boldsymbol{\phi}_t, \mathbf{\Theta}_t\}_{t=1}^T)$ \Comment{Compute physics loss}
        \State $\mathbf{r}, \boldsymbol{\phi}, \mathbf{z} \leftarrow \arg\min( 
        \frac{1}{T}\sum_{t=1}^T \mathcal{L}_{\text{fit}}^t
        + \lambda_{\text{physics}} \mathcal{L}_{\text{physics}})$
        \For{$t = 1$ to $T$}
            \State $\mathbf{\Theta}_t \leftarrow \mathcal{D}(\mathbf{z}_t)$ \Comment{Decode VPoser latent}
            \State $\mathcal{M}_t \leftarrow (\mathbf{r}_t, \boldsymbol{\phi}_t, \mathbf{\Theta}_t)$ \Comment{Update human pose}
        \EndFor
        \State\textbf{return} $\{(\mathcal{M}_t, \mathbf{P}_t)\}_{t=1}^T$
    \EndFunction
    
    \State $\{\mathbf{I}_t,\mathbf{M}_\mathcal{H}^t,\mathbf{M}_\mathcal{O}^t\}_{t=0}^T \leftarrow \text{GenerateHSIVideo}()$
    \State $\tau \leftarrow \text{Reconstruct4DHSI}()$
    \State $\tau \leftarrow \text{Refinement}()$
    
    \State\textbf{Output:} 4D HSI sequence $\tau = \{(\mathcal{M}_t, \mathbf{P}_t)\}_{t=1}^T$
    \end{algorithmic}
\end{algorithm*}

\end{document}